\documentclass{ecai} 

\usepackage{color}
\usepackage{amsthm}
\usepackage{xcolor}
\usepackage{amsmath}
\usepackage{amssymb}
\usepackage{amsfonts}
\usepackage{booktabs}
\usepackage{enumitem}
\usepackage{graphicx}
\usepackage{latexsym}
\usepackage{multirow} 
\usepackage{algorithm} 
\usepackage{algorithmic}

\newcommand{\eg}{\textit{e}.\textit{g}.}

\begin{document}

\begin{frontmatter}

\paperid{6830} 

\title{Diffusion Noise Feature: \\Accurate and Fast Generated Image Detection}

\author[A]{\fnms{Yichi}~\snm{Zhang}}
\author[A]{\fnms{Xiaogang}~\snm{Xu}\thanks{Corresponding Author. Email: xiaogangxu00@gmail.com.}} 

\address[A]{Zhejiang University}

\begin{abstract}
Generative models now produce images with such stunning realism that they can easily deceive the human eye. While this progress unlocks vast creative potential, it also presents significant risks, such as the spread of misinformation. 
Consequently, detecting generated images has become a critical research challenge. However, current detection methods are often plagued by low accuracy and poor generalization.
In this paper, to address these limitations and enhance the detection of generated images, we propose a novel representation, \textsc{\underline{D}iffusion \underline{N}oise \underline{F}eature} (DNF). 
Derived from the inverse process of diffusion models, DNF effectively amplifies the subtle, high-frequency artifacts that act as fingerprints of artificial generation. Our key insight is that real and generated images exhibit distinct DNF signatures, providing a robust basis for differentiation.
By training a simple classifier such as ResNet-50 on DNF, our approach achieves remarkable accuracy, robustness, and generalization in detecting generated images, including those from unseen generators or with novel content.
Extensive experiments across four training datasets and five test sets confirm that DNF establishes a new state-of-the-art in generated image detection.
The code is available at https://github.com/YichiCS/Diffusion-Noise-Feature.

\end{abstract}

\end{frontmatter}


\section{Introduction}\label{sec:intro}

\begin{figure*}[t]
\label{fig:overview}
\centering
\includegraphics[width=\linewidth]{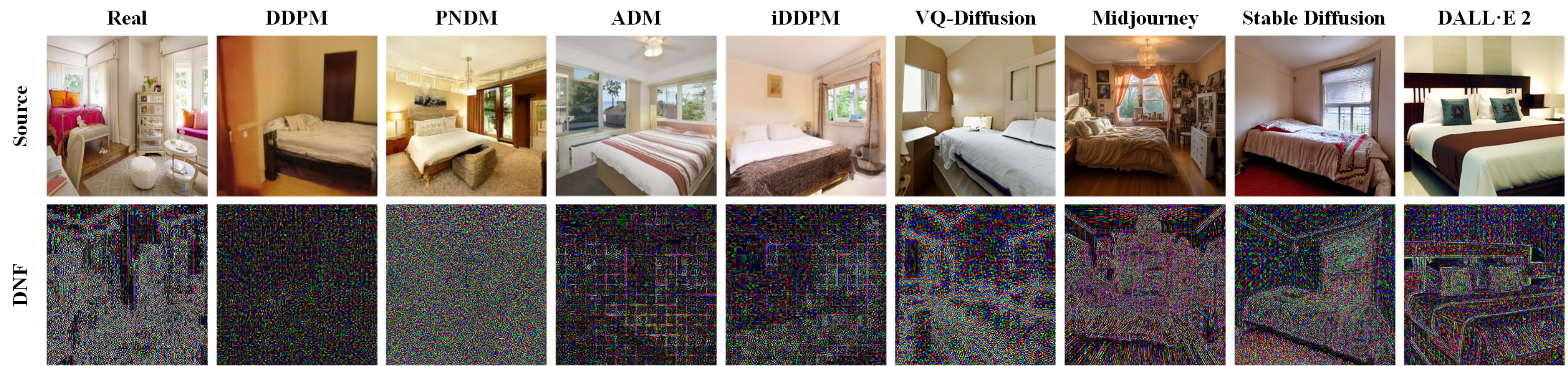}
\caption{Visualizations of DNF on real and generated images from eight generators.}
\end{figure*}

In recent years, deep generative models have revolutionized visual content synthesis through several seminal frameworks: variational autoencoders (VAEs)~\cite{kingma2013auto}, generative adversarial networks (GANs)~\cite{goodfellow2020generative}, and diffusion models (DMs)~\cite{ho2020denoising}. These architectures have fundamentally reshaped the landscape of image generation through distinct yet complementary approaches. The VAE framework established a principled probabilistic foundation for latent representation learning, while GAN variants~\cite{brock2018large,karras2017progressive,karras2019style,zhu2017unpaired,choi2018stargan} demonstrated remarkable quality in adversarial synthesis through innovative network designs and training strategies. Most recently, diffusion-based approaches~\cite{dhariwal2021diffusion,rombach2022high} have emerged as the state-of-the-art paradigm, achieving unprecedented photorealism through progressive denoising processes. This evolution of generative architectures has not only expanded the theoretical understanding of data distributions but also enabled practical applications ranging from artistic creation to visual content augmentation.

The unprecedented capabilities of modern generative models have simultaneously raised critical ethical dilemmas and societal safety concerns. The proliferation of multimodal architectures now enables open-ended text-to-image synthesis through natural language prompts, effectively democratizing visual content creation. Particularly concerning is the emergence of DeepFake technologies~\cite{westerlund2019emergence}, which achieves imperceptible facial identity replacement not only in recorded videos but also real-time video streams. Malicious actors can weaponize these generative capabilities to orchestrate sophisticated attacks spanning privacy violations~\cite{murdoch2021privacy}, financial extortion~\cite{uyyala2023advanced}, and identity theft through synthetic media impersonation~\cite{qi2019exploiting,giglietto2019fake}. These social problems create a pressing need for reliable forensic methods to distinguish synthetic from authentic media. Such methods must not only detect subtle generation artifacts but also evolve alongside sophisticated new algorithms, thus safeguarding digital trust in the generative AI era.

Significant research efforts have been dedicated to developing robust frameworks for detecting AI-generated visual content. Current methodologies rely on analyzing post-processing artifacts~\cite{wang2020cnn}, examining spectral discrepancies in the frequency domain~\cite{qian2020thinking}, and employing reconstruction-based forensic techniques~\cite{wang2023dire}. These approaches have proven effective at distinguishing synthetic outputs from conventional generative models such as VAEs and GANs~\cite{shi2023discrepancy,tan2023learning}, while also capturing manipulation traces from image enhancement pipelines like deblurring~\cite{zhong2023rich}, denoising~\cite{frank2020leveraging}, super-resolution, and facial manipulation~\cite{sinitsa2023deep,rana2022deepfake}.
However, the proliferation of diffusion models poses an unprecedented challenge to detection. Modern text-to-image systems like DALL·E~\cite{betker2023improving}, Stable Diffusion~\cite{rombach2022high} (SD), and Midjourney~\cite{midjourney2022} produce more photorealistic outputs with fewer spectral artifacts than prior models. This realism cripples traditional artifact-based detectors, causing critical failures in accuracy and generalization, especially for high-resolution images. 

To address these challenges, we introduce the \textsc{\underline{D}}iffusion \underline{N}oise \underline{F}eature (DNF), a novel representation designed to capture the intrinsic distinctions between real and generated images. Unlike conventional approaches that directly analyze pixel or frequency domains, our method extracts DNF from the estimated noise sequence generated by a pretrained DM during the inverse diffusion process. Training a classifier on DNF easily yields a generated image detector with improved detection performance, generalization, and robustness.

Our method is inspired by the unique training and inference mechanisms of DMs. Through the iterative process of adding and removing noise from images, DMs develop exceptional sensitivity to high-frequency image components. 
This property manifests distinctively during the inverse diffusion process, where the inverse diffusion process in a pre-trained DM can disrupt high-frequency details in the image by adding estimated noise, transforming them into a unified noise distribution. 
Crucially, the intrinsic distributional differences between real and generated images become amplified in these estimated noises, as evidenced in Figure~\ref{fig:overview}. The estimated noise from real images typically exhibits chaotic patterns, while that from generated images demonstrates characteristic high-frequency details, grid-like artifacts, or residual high-frequency components from the original image. This distinctive phenomenon enables the construction of effective estimated-noise-based representations for discriminating generated images. 
More importantly, we observe that this phenomenon is not limited to images generated by DMs, but is prevalent across all types of generated images. In other words, DMs can accurately capture the generative fingerprints of an image by perturbing its high-frequency details in the inverse diffusion process.

To operationalize this insight, we employ a pre-trained and converged DM to process input images for detection. We utilize this DM to apply the inverse diffusion process to the image, progressively transforming it into pure Gaussian noise while collecting estimated noises generated at each step. Subsequently, we employ a tailored feature fusion strategy to construct DNF from the noise sequence. This method effectively channels the DM's inherent sensitivity to high-frequency image information into the classifier trained on DNF, ensuring the detection's accuracy, generalization, and robustness.

Extensive experiments across four training datasets and five test benchmarks validate the state-of-the-art performance of the classifier trained on DNF in generated image detection:
(1) Our DNF classifier achieves \textbf{99.8\% accuracy} in evaluation across the five test datasets, significantly outperforming the 87.7\% average of all baseline methods~\cite{wang2020cnn,frank2020leveraging,chai2020makes,shiohara2022detecting,wang2023dire} compared to their 87.7\% mean accuracy. 
(2) The classifier maintains \textbf{remarkable robustness} against common image perturbations, retaining 99.2\% accuracy under Gaussian blur or JPEG compression, which represents a huge improvement over the best-performing baseline.
(3) The DNF classifier exhibits \textbf{strong cross-dataset and cross-generator generalization capabilities}, achieving high accuracy on detecting images from different datasets (\eg, LSUN-Bedroom~\cite{yu2015lsun}, ImageNet~\cite{deng2009imagenet} or CelebA~\cite{liu2018large}) and across various generators (\eg, GANs, DMs), even when trained on only a limited variety of generators. 

In summary, our main contributions are as follows:
\begin{itemize}
    \item We introduce a novel image representation, \textbf{DNF}, which pioneers the use of estimated noise from the inverse diffusion process to construct image representations for generated image detection.
    \item We conducted comprehensive experiments to prove classifier trained on DNF achieves state-of-the-art performance in generated image detection, significantly outperforming existing baselines.
    \item We develop a rigorous evaluation methodology for real-world detection scenarios, explicitly incorporating assessment of cross-domain generalization across diverse generators and datasets.
\end{itemize}

\section{Related Work}\label{sec2}

\subsection{Generative Models: GANs and DMs}

GANs~\cite{goodfellow2020generative} establish an unsupervised adversarial learning paradigm that progressively refines synthesized images toward photorealism through discriminator-guided generator optimization.
Unconditional architectures exemplified by BigGAN~\cite{brock2018large} and StyleGAN series~\cite{karras2019style,karras2020analyzing,karras2021alias} model high-dimensional latent distributions of authentic data, enabling the generation of high-fidelity visual content. 
Conditional variants extend this paradigm by incorporating cross-modal constraints. Representative models like CycleGAN~\cite{zhu2017unpaired} and StarGAN~\cite{choi2018stargan} explicitly design cyclic consistency losses and domain adaptation modules, respectively, achieving state-of-the-art performance in image-to-image translation tasks.

DMs~\cite{gu2022vector,phung2023wavelet,li2023q,peebles2023scalable} operate through iterative denoising processes that progressively transform random noise into coherent images, establishing new frontiers in generative modeling.
DDPM~\cite{ho2020denoising} uses a forward-backward Markov chain to progressively corrupt images with Gaussian noise and then reverse the process via conditional denoising, thereby optimizing a variational lower bound on the data likelihood.
DDIM~\cite{song2020denoising} reparameterizes the diffusion trajectory as a non-Markovian process to enable deterministic sampling via learned ODEs, preserving generation quality while drastically reducing inference steps.
Text-to-image DMs like Stable Diffusions~\cite{rombach2022high} and Midjourney~\cite{midjourney2022} achieve remarkable image generation by performing diffusion in latent space and training on large-scale datasets.

\subsection{Generated Image Detection}

To mitigate potential risks associated with generated images, researchers are gradually paying attention to generated image detection~\cite{sinitsa2023deep,tan2023learning,chai2020makes,corvi2023detection,cao2022end}. 
CNNDetection~\cite{wang2020cnn} has discovered artifacts in the frequency domain of CNN-generated images, making detection of generated images feasible. It has constructed the first universal CNN-generated image detector through post-processing of images. 
Similarly, FrequencyDetection~\cite{qian2020thinking} classifies generated and real images by observing features presented after discrete cosine transformation. 
DisGRL~\cite{shi2023discrepancy} incorporates three proposed components to learn both forgery-sensitive and genuine compact visual patterns. 
DIRE~\cite{wang2023dire} utilizes the DM to reconstruct images and observes the differences between the original and reconstructed images for image detection. 
LaRE$^2$~\cite{luo2024lare} performs detection by reconstructing the image error in the latent space.
DRCT~\cite{chendrct} proposes utilizing high-quality diffusion reconstruction to generate hard samples and employs contrastive training to guide the learning of diffusion artifacts.

\section{Method}
\label{sec:method}

In this section, we introduce \textsc{\underline{D}iffusion \underline{N}oise \underline{F}eature}, and present the motivation behind its design, the process of feature extraction, and its application to generated image detection.

\subsection{Motivation}

Generative models aim to learn a model distribution $p_{\text{gen}}(\mathbf{x})$ that accurately approximates the data distribution $p_{\text{data}}(\mathbf{x})$. When the statistical discrepancy between these distributions falls below perceptibility thresholds, synthetic samples become indistinguishable from real images to both human observers and automated detectors. Formally, this condition can be characterized by the KL divergence criterion $D_{\text{KL}}(p_{\text{data}} \parallel p_{\text{gen}}) < \epsilon$, where $\epsilon$ denotes the perceptual indistinguishability bound.
Our objective is to find a discriminative feature mapping $\boldsymbol{f}: \mathcal{X} \to \mathcal{Z}$ that induces maximally separable distributions in the new feature space. Let $q_{\text{data}}(\mathbf{z})$ and $q_{\text{gen}}(\mathbf{z})$ denote the feature-space distributions of real and generated images respectively. The transformation should satisfy $D_{\text{KL}}(q_{\text{data}} \parallel q_{\text{gen}}) \geq \delta \gg \epsilon$, where $\delta$ represents an amplified discriminability threshold.

We naturally consider employing DMs to disentangle the two data distributions $p_{\text{gen}}(\mathbf{x})$ and $p_{\text{data}}(\mathbf{x})$, primarily motivated by two insights: 
(1) While existing generative models exhibit remarkable accuracy in synthesizing low-frequency characteristics that closely resemble real images, they still struggle to capture high-frequency patterns. Assuming our target mapping $\boldsymbol{f}$ operates as a high-pass filter to preserve high-frequency components, we observe its partial effectiveness in distribution separation, though insufficient to guarantee $\delta \gg \epsilon$. Conversely, DMs inherently perform iterative denoising to progressively reconstruct images from noise, making them particularly adept at capturing and amplifying fine-grained details through this multi-stage refinement process. 
(2) Generative architectures with different inductive biases tend to converge to similar image distributions under identical pretraining protocols. This similarity makes DMs more effective at disrupting high-frequency artifacts in synthetic images while posing greater challenges when altering high-frequency content in real images. Consequently, the estimated noise distributions for generated versus authentic images exhibit distinct characteristics, as visualized in Figure~\ref{fig:overview}, reflecting their fundamentally different responses to the diffusion process.

Overall, our method consists of three stages: \texttt{Estimated Noise Extraction}, \texttt{Feature Fusion}, and \texttt{Classifier Training}. In the following sections, we will provide detailed implementation and design principles.

\subsection{Estimated Noise Extraction}

In this stage, we aim to obtain the estimated noise sequence $\{\epsilon_i\}$ from the input image $\boldsymbol{x}$. Within the framework of DDIM~\cite{song2020denoising}, $\boldsymbol{x}_{t-1}$ can be sampled from $\boldsymbol{x}_t$ via:
\begin{align}
\boldsymbol{x}_{t-1} &= \sqrt{\alpha_{t-1}}\left(\frac{\boldsymbol{x}_t - \sqrt{1-\alpha_t} \, \epsilon_\theta^{(t)}(\boldsymbol{x}_t)}{\sqrt{\alpha_t}}\right)  \nonumber\\
&+ \sqrt{1 - \alpha_{t-1} - \sigma_t^2} \, \epsilon_\theta^{(t)}(\boldsymbol{x}_t) 
+ \sigma_t \epsilon_t.
\label{eq:ddim_forward}
\end{align}
where $\epsilon_t\!\sim\!\mathcal{N}(0, \mathbf{I})$ represents Gaussian noise independent of $\boldsymbol{x}$, $\epsilon_\theta^{(t)}(\boldsymbol{x}_t)$ represents the estimated noise output by the DM $\mathcal{N}_\text{diff}$ with parameter $\theta$. $t$ denotes the time step, $\alpha$ is a hyperparameter, and $\sigma_t$ controls the diffusion process. Specifically, when $\sigma_t = \sqrt{(1-\alpha_{t-1})/(1 - \alpha_t)}\sqrt{1 - \alpha_t / \alpha_{t-1}}$, the process corresponds to the sampling process in DDPM~\cite{ho2020denoising}.
In DDIM, setting $\sigma_t = 0$ allows the process to be determined by $\boldsymbol{x}_{t}$ and $\boldsymbol{x}_0$. Thus, the Equation~\ref{eq:ddim_forward} can be rewritten as:
\begin{align}
\frac{\boldsymbol{x}_{t-1}}{\sqrt{\alpha_{t-1}}} = \frac{\boldsymbol{x}_{t}}{\sqrt{\alpha_{t}}} + \left(\sqrt{\frac{1-\alpha_{t-1}}{\alpha_{t-1}}} - \sqrt{\frac{1-\alpha_{t}}{\alpha_{t}}}\right)\epsilon_{\theta}^{(t)}(\boldsymbol{x}_t).
\label{eq:ddim_forward_sigma_0}
\end{align}

We need to reverse this process, which differs from how DMs are typically applied in practice. Next, we will demonstrate that restoring an image to pure noise through the inverse diffusion process is reasonable.
Typically, when the total diffusion steps $T$ of a DM is a sufficiently large number, such as 1000, Equation~\ref{eq:ddim_forward_sigma_0} can be interpreted as an Euler method for integrating ordinary differential aligns (ODEs). Therefore, it can be rewritten as:
\begin{align}
    \frac{\boldsymbol{x}_{t-\Delta t}}{\sqrt{\alpha_{t-\Delta t}}} = \frac{\boldsymbol{x}_{t}}{\sqrt{\alpha_{t}}} + \left(\sqrt{\frac{1-\alpha_{t-\Delta t}}{\alpha_{t-\Delta t}}} - \sqrt{\frac{1-\alpha_{t}}{\alpha_{t}}}\right)\epsilon_{\theta}^t(\boldsymbol{x}_t).
\end{align}
%
Defining $\lambda_t\!=\!\sqrt{(1\!-\!\alpha_t) / \alpha_t}$, $\boldsymbol{y}_t\!=\!\boldsymbol{x}_t / \sqrt{\alpha}_t$, the corresponding ODE is then reformulated as: 
\begin{align}
    \text{d}\boldsymbol{y}_t &= \epsilon_{\theta}^{(t)}(\frac{\boldsymbol{y}_t}{\sqrt{\lambda_t^2+1}})\text{d}\lambda_t.
\end{align}
Thus, we can reverse the align and execute the reverse diffusion process starting from $ t = 0 $, that is 
\begin{align}
    \frac{\boldsymbol{x}_{t+1}}{\sqrt{\alpha_{t+1}}} = \frac{\boldsymbol{x}_{t}}{\sqrt{\alpha_{t}}} + \left(\sqrt{\frac{1-\alpha_{t+1}}{\alpha_{t+1}}} - \sqrt{\frac{1-\alpha_{t}}{\alpha_{t}}}\right)\epsilon_{\theta}^{(t)}(\boldsymbol{x}_t).
\label{eq:inverse_diffusion_process_tau}
\end{align}

Specifically, to accelerate this process, we select a subsequence $\{\tau_i\}$ from the time steps $\{0, 1, ..., T-1\}$. Therefore, the final inverse diffusion process can be written as: 
\begin{align}
    \frac{\boldsymbol{x}_{\tau_{i+1}}}{\sqrt{\alpha_{\tau_{i+1}}}}\!=\!\frac{\boldsymbol{x}_{\tau_{i}}}{\sqrt{\alpha_{\tau_{i}}}}\!+\!\left(\sqrt{\frac{1\!-\!\alpha_{\tau_{i+1}}}{\alpha_{\tau_{i+1}}}}-\sqrt{\frac{1\!-\!\alpha_{\tau_{i}}}{\alpha_{\tau_{i}}}}\right)\epsilon_{\theta}^{(\tau_{i})}(\boldsymbol{x}_{\tau_{i}}).
\end{align}

In the actual implementation, we first train the DM $\mathcal{N}_{\text{diff}}$ to convergence with parameters $\theta$ on the dataset $\mathcal{D}_\text{pre}$. Then, for image $\boldsymbol{x}$ to be detected, we repeatedly apply the inverse diffusion process as described in Equation~\ref{eq:inverse_diffusion_process_tau}. We collect the estimated noise $\epsilon_\theta^{(\tau_{i})}(\boldsymbol{x}_{\tau_{i}})$ generated at each step $\tau_{i}$ and integrate it into an estimated noise sequence $\{\epsilon_i\}$. We denote this algorithm as $\boldsymbol{f}: \boldsymbol{x} \rightarrow \{\epsilon_i\}$. The detailed algorithm is presented in Algorithm~\ref{alg:obtain_ens}.

\begin{algorithm}[tb]
    \caption{Obtaining Estimated Noise Sequence $f$}
    \label{alg:obtain_ens}
    \textbf{Input}: Image $\boldsymbol{x}$, DM $\mathcal{N}_\text{diff}$, Dataset $\mathcal{D}_\text{pre}$\\
    \textbf{Parameter}: Step $T$, hyper-parameter $\alpha$, sub-Step $\{\tau_i\}$\\
    \textbf{Output}: Noise Sequence $\{\epsilon_i\}$
    \begin{algorithmic}[1] 
        \STATE $\theta = \texttt{PreTrain}(\mathcal{N}_\text{diff}, \mathcal{D}_\text{pre})$
        \STATE $\texttt{EN-Seq} \leftarrow \{\}, \boldsymbol{x}_{\tau_0} \leftarrow \boldsymbol{x}$
        \FOR{$\tau_i \in \{\tau_i\}$}
        
        \STATE $\epsilon_\theta^{(\tau_i)}(\boldsymbol{x}_{\tau_i}) \leftarrow \mathcal{N}_\text{diff}(\boldsymbol{x}_{\tau_i}, \tau_i; \theta)$
        
        \STATE $\texttt{EN-Seq} \leftarrow \texttt{EN-Seq} + \epsilon_\theta^{(\tau_i)}(\boldsymbol{x}_{\tau_i})$
        
        \STATE $\hat{\boldsymbol{x}}_0 \leftarrow (\boldsymbol{x}_{\tau_i} - \sqrt{1-\alpha_{\tau_i}} \, \epsilon_\theta^{({\tau_i})}(\boldsymbol{x}_{\tau_i})) / \sqrt{\alpha_{\tau_i}}$
        
        \STATE $\boldsymbol{x}_{\tau_{i+1}} = \sqrt{\alpha_{\tau_i}}\hat{\boldsymbol{x}}_0 + \sqrt{1 - \alpha_{\tau_i}} \, \epsilon_\theta^{({\tau_i})}(\boldsymbol{x}_{\tau_i})$
        
        \ENDFOR
        \STATE $\{\epsilon_i\} \leftarrow \texttt{EN-Seq}$
    \end{algorithmic}
\end{algorithm}

\begin{table*}[t]
\caption{Generated image detection performance of DNF and baselines on the LSUN-Bedroom split of the DiffusionForensics.} 
\centering
\resizebox{\linewidth}{!}{
\begin{tabular}{@{}lccccccccccc@{}}
    \toprule
    \multirow{2}{*}{Method} & \multicolumn{10}{c}{Testing Generators} & Total  \\ 
    & ADM$^\dag$ & DDPM & iDDPM$^\dag$ & LDM & PNDM$^\dag$ & SD-v2 & VQ-D & DALL-E 2 & IF & Midjourney & Avg. \\
    \midrule
    CNNDet.     & 50.1/63.5 & 50.2/79.4 & 50.2/78.0 & 50.1/61.4 & 50.1/60.3 & 50.8/80.7 & 50.1/70.8 & 52.8/87.4 & 51.3/79.9 & 50.9/58.5 & 50.6/71.9 \\
    Patchfor.    & 50.2/67.4 & 53.2/74.2 & 51.2/63.4 & 56.7/89.1 & 56.5/72.4 & 54.2/72.7 & 87.2/95.4 & 50.1/68.9 & 50.0/56.3 & 56.1/57.2 & 56.5/71.7 \\
    SBI         & 53.4/60.8 & 56.9/50.8 & 58.4/56.2 & 83.4/90.2 & 73.1/95.6 & 59.2/70.9 & 56.2/74.2 & 51.2/56.4 & 61.3/72.3 & 52.3/87.9 & 60.5/71.5 \\
    DIRE        & 94.7/99.7 & 92.6/99.6 & 94.6/99.7 & 94.6/99.5 & 94.3/99.1 & 94.6/99.7 & 94.6/99.8 & 89.5/99.5 & 94.6/99.7 & 92.1/98.0 & 93.6/99.4 \\
    NPR         & 88.7/99.1 & 99.9/\textbf{100} & 91.2/99.7 & \textbf{100}/\textbf{100} & 92.3/\textbf{100} & 94.5/\textbf{100} & 96.3/98.2 & 94.4/99.8 & 83.2/97.8 & 82.1/98.1 & 92.3/99.3 \\
    LaRE$^2$    & 98.4/99.6 & 92.3/98.4 & 96.5/99.4 & 86.4/92.1 & 95.4/98.9 & 89.9/98.2 & 91.2/99.2 & 84.4/90.5 & 72.2/93.6 & 76.3/88.2 & 88.3/95.8 \\
    FatFormer   & 98.3/99.8 & 91.2/98.2 & 95.2/99.4 & 96.6/99.1 & 94.5/99.2 & 93.2/98.6 & 99.1/\textbf{100} & 78.6/88.2 & 82.1/90.3 & 72.6/85.5 & 90.1/95.8 \\
    F3Net$^*$      & 91.2/97.8 & 90.7/98.5 & 89.9/99.2 & 98.1/\textbf{100}  & 92.3/97.2 & 81.1/90.4 & 92.4/97.3 & 78.1/86.2 & 73.6/82.2 & 75.9/81.1 & 86.3/92.9 \\
    Patchfor.$^*$   & 94.1/99.8 & 72.9/98.2 & 95.2/99.4 & 97.2/\textbf{100} & 94.2/\textbf{100} & 74.5/90.2 & 95.4/100  & 85.2/98.2 & 65.4/82.3 & 53.2/88.6 & 83.7/95.7 \\
    CNNDet.$^*$     & 98.8/99.9 & 98.5/99.9 & 99.1/99.9 & 97.9/99.8 & 99.1/99.9 & 80.4/93.5 & 78.8/94.6 & 94.5/98.5 & 80.3/94.0 & 53.4/58.1 & 88.1/93.8 \\

    DNF (Ours)  & \textbf{100}/\textbf{100} & \textbf{99.7}/\textbf{100} & \textbf{100}/\textbf{100} & \textbf{100}/\textbf{100} & \textbf{100}/\textbf{100} & \textbf{100}/\textbf{100} & \textbf{99.8}/\textbf{100} & \textbf{100}/\textbf{100} & \textbf{99.9}/\textbf{100} & \textbf{98.9}/\textbf{99.9} & \textbf{99.8}/\textbf{99.9} \\ 
    \bottomrule
\end{tabular}}
\label{tab:baseline}
\end{table*}

\subsection{Feature Fusion}

\texttt{Feature fusion} serves as a pivotal step in transforming the estimated noise sequence $\mathcal{E} = \{\epsilon_t\}_{t=0}^{N-1} \in \mathbb{R}^{N \times H \times W \times C}$ into an image representation $\hat{\boldsymbol{x}} \in \mathbb{R}^{1 \times H \times W \times C}$ for training generated image classifiers. This process aims to seek a fusion operation $\boldsymbol{g}:\mathcal{E} \to \hat{\boldsymbol{x}}$ while preserving dimensional consistency with the original images $\boldsymbol{x}$. The transformation needs to carefully handle both changes over time in the estimated noise and the preservation of important visual details, which means combining information from different time steps while keeping the overall image structure intact.

\texttt{First Noise Strategy} directly utilizes the initial estimated noise $\epsilon_\theta^{({\tau_0})}(\boldsymbol{x})$ as the feature representation. Since $\epsilon_\theta^{({\tau_0})}(\boldsymbol{x})$ is derived directly from the original image $\boldsymbol{x}$ and serves as a direct estimation of its high-frequency details, it effectively encodes critical visual cues essential for generated image detection.  
\begin{align}  
\hat{\boldsymbol{x}} = \boldsymbol{g}(\mathcal{E}) = \mathcal{E}[0, :, :, :] = \epsilon_0  
\end{align}  
However, this approach may neglect complementary information contained in subsequent steps of the inverse diffusion process.

\texttt{Mean Noise Strategy} comprehensively leverages information across the entire inverse diffusion process by computing temporal averaging over the estimated noise sequence $\mathcal{E}$. During the inverse diffusion process, the DM progressively transforms the data distribution $p(\boldsymbol{x}_t)$ toward a standard normal prior. Each noise estimate $\epsilon_t$ captures the distinct distributional discrepancy between the current state $\boldsymbol{x}_t$ and the target distribution $p(\boldsymbol{x}_{t-1}|\boldsymbol{x}_t)$.  
\begin{align}  
\hat{\boldsymbol{x}} = \boldsymbol{g}(\mathcal{E}) = \frac{1}{T} \sum_{t=0}^{T-1} \mathcal{E}[t, :, :, :]  
\end{align}  
Since the temporal aggregation operation captures the divergence between the data distribution and the target prior, $\hat{\boldsymbol{x}}$ becomes a powerful representation for detecting generated images.

\texttt{Convolutional Strategy} introduces a learnable transformation via 3D convolution $\texttt{Conv3D}(\boldsymbol{\mathcal{E}}; \boldsymbol{\Theta})$ to achieve adaptive fusion of spatiotemporal patterns in noise sequences. We connect this convolutional layer before the classifier and train it end-to-end. 
\begin{align}  
\hat{\boldsymbol{x}} = \boldsymbol{g}(\boldsymbol{\mathcal{E}}; \boldsymbol{\Theta}) = \texttt{Conv3D}(\boldsymbol{\mathcal{E}}; \boldsymbol{\Theta}),  
\end{align}  
where the convolutional kernel $\boldsymbol{\Theta}$ operates across temporal, spatial, and channel dimensions to enable joint modeling of evolving patterns in the estimated noise sequence.

The selection of the three fusion strategies is driven by the trade-off between computational efficiency and detection accuracy. We evaluate their performance through extensive experiments to determine the optimal strategy under varying scenarios.

\subsection{Classifier Training \& Inference}

Following the estimated noise sequence extraction $\boldsymbol{f}$ and feature fusion $\boldsymbol{g}$, the input image $\boldsymbol{x}$ is progressively transformed into its DNF $\hat{\boldsymbol{x}}$, retaining identical spatial dimensions ($H \times W \times C$) to the original input. This transformation is formally defined as:  
\begin{align}  
\boldsymbol{x} \xrightarrow[\mathcal{N}_{\text{diff}}, \mathcal{D}_{\text{pre}}]{\boldsymbol{g} \circ \boldsymbol{f}} \hat{\boldsymbol{x}},  
\end{align}  
where $\mathcal{N}_{\text{diff}}$ denotes the architecture of the pre-trained diffusion model, and $\mathcal{D}_{\text{pre}}$ represents the dataset used for its pretraining. The composite operation $\boldsymbol{g} \circ \boldsymbol{f}$ first extracts the noise sequence $\boldsymbol{\mathcal{E}} = \{\epsilon_t\}_{t=0}^{N-1}$ through $\boldsymbol{f}$, then fuses it into $\hat{\boldsymbol{x}}$ via $\boldsymbol{g}$.  
Let $\mathcal{N}{\text{c}}$ denote the generated image classifier. The end-to-end inference pipeline in Equation~\ref{eq:pipeline}integrates both DNF extraction and classification stages, where the output $\{0, 1\}$ indicates whether $\boldsymbol{x}$ is a generated image.
\begin{align}\label{eq:pipeline}
\mathcal{N}(\boldsymbol{x}) = \mathcal{N}_{\text{c}}\left(\boldsymbol{g} \circ \boldsymbol{f}(\boldsymbol{x}; \mathcal{N}_{\text{diff}})\right) \to \{0, 1\}.
\end{align}

\section{Experiments}\label{sec:exp}

In this section, we provide extensive experiments to demonstrate that our method outperforms other baselines in terms of generated image detection performance. Our experiments include baseline evaluations, generalization evaluation (which cover both model and dataset generalization), and perturbation robustness evaluation. Additionally, we conduct ablation studies to investigate the impact of various components on the effectiveness of our approach.

\begin{table*}[t]
\centering
\caption{Generalization performance of DNF, DIRE, and CNNDetection across four training sets and five test sets}
\resizebox{\linewidth}{!}{
\begin{tabular}{@{}llccccccccccc@{}}
    \toprule
    \multirow{2}{*}{Method} & Training & \multicolumn{2}{c}{DF ImageNet}& \multicolumn{4}{c}{GenImage} & \multicolumn{4}{c}{DF CelebA} & Total  \\ 
    & Dataset   & ADM$^\dag$      & SD-v1     & SD-v1.4   & SD-v1.5   & Glide     & wukong    &  SD-v2$^\dag$   & Mid.      & DALL-E 2  & IF        & Avg.      \\
    \midrule
    \multirow{4}{*}{CNNDet.}             & LSUN-B.   & 63.6/80.6 & 53.3/63.8 & 52.8/55.0 & 53.0/56.0 & 78.3/88.1 & 50.8/51.8 & 12.9/9.8  & 11.8/7.7  & 49.0/49.4 & 12.8/9.6  & 43.8/47.2      \\
    & ImageNet  & 71.6/79.8 & 51.0/51.2 & 41.3/40.9 & 40.6/40.5 & 60.5/63.4 & 45.9/48.9 & 37.0/41.6 & 48.4/49.1 & 54.2/52.2 & 36.5/41.2 & 48.7/50.9      \\
    & CelebA    & 51.0/58.8 & 52.6/68.0 & 51.1/50.3 & 52.9/57.5 & 50.5/50.0 & 53.1/57.1 & 78.4/69.9 & 73.6/67.7 & 54.2/52.2 & 53.6/53.9 & 57.1/58.5      \\    
    & CNNSpot   & 51.2/82.0 & 50.5/69.5 & 50.4/59.4 & 50.6/60.1 & 52.4/68.6 & 50.6/59.0 & 52.8/87.4 & 54.9/90.1 & 53.8/87.9 & 50.3/61.3 & 51.8/72.5      \\
    \midrule
    \multirow{4}{*}{DIRE}               & LSUN-B.   & 99.8/99.8 & \textbf{99.1}/99.9 & 91.2/98.6 & 91.6/98.8 & 92.4/99.5 & 90.1/98.3 & 49.9/49.9 & 50.4/50.2 & 50.4/50.2 & 50.3/50.2 & 76.5/79.5      \\
    & ImageNet  & 99.8/99.9 & 98.2/99.9 & 95.4/99.7 & 96.3/99.9 & 67.2/73.1 & 52.8/63.8 & 50.0/50.0 & 50.0/50.0 & 50.0/50.0 & 50.0/50.0 & 71.0/73.6      \\
    & CelebA    & 99.8/99.9 & 58.2/66.2 & 53.4/62.1 & 55.8/67.8 & 63.1/71.5 & 66.8/78.8 & 96.7/\textbf{100}  & 95.0/\textbf{100} & 93.4/\textbf{100} & 96.8/\textbf{100} & 77.9/84.6     \\
    & CNNSpot   & 72.8/83.4 & 50.1/50.1 & 51.2/53.6 & 49.8/50.1 & 73.4/76.8 & 58.6/61.2 & 50.1/50.2 & 58.2/62.9 & 67.2/75.3 & 52.1/53.3 & 58.4/61.7     \\
    
    \midrule
    \multirow{4}{*}{DNF (Ours)}                & LSUN-B.   & 98.0/\textbf{100} & 96.3/\textbf{100} & 98.6/99.9 & 98.6/99.9 & 99.9/\textbf{100} & 99.7/\textbf{100} & 75.5/99.8 & 97.5/99.9 & \textbf{100}/\textbf{100} & \textbf{100}/\textbf{100} & 96.4/\textbf{99.9} \\ 
    & ImageNet  & \textbf{100}/\textbf{100} & 98.9/99.9 & \textbf{100}/\textbf{100} & \textbf{100}/\textbf{100} & \textbf{100}/\textbf{100} & \textbf{100}/\textbf{100} & 98.7/\textbf{100} & 99.0/\textbf{100} & \textbf{100}/\textbf{100} & \textbf{100}/\textbf{100} & \textbf{99.9}/\textbf{99.9} \\ 
    & CelebA    & \textbf{100}/\textbf{100} & 98.9/\textbf{100} & 99.7/99.9 & 99.8/\textbf{100} & 99.7/99.9 & 99.8/\textbf{100} & \textbf{100}/\textbf{100} & \textbf{100}/\textbf{100} & \textbf{100}/\textbf{100} & \textbf{100}/\textbf{100} & 99.7/\textbf{99.9} \\ 
    & CNNSpot   & 86.9/\textbf{100} & 77.7/\textbf{100} & 77.5/99.1 & 77.8/99.1 & 79.2/96.6 & 80.3/98.7 & 60.6/99.1 & 86.1/99.7 & 85.0/99.7 & 75.8/99.6 & 78.7/99.5 \\ 
    \toprule
    \multirow{2}{*}{Method} & Training & \multicolumn{10}{c}{DF LSUN-Bedroom} & Total  \\ 
    & Dataset   & ADM$^\dag$      & DDPM      &  iDDPM$^\dag$   & LDM       & PNDM$^\dag$     & SD-v2     & VQ-D      & DALL-E 2  & IF        & Mid.      & Avg.      \\
    \midrule
    \multirow{4}{*}{CNNDet.}             & LSUN-B.   & 98.8/99.9 & 98.5/99.9 & 99.1/99.9 & 97.9/99.8 & 99.1/99.9 & 80.4/93.5 & 78.8/94.6 & 94.5/98.5 & 80.3/94.0 & 53.4/58.1 & 88.1/93.8 \\
    & ImageNet  & 72.4/74.1 & 71.2/65.7 & 76.8/80.8 & 64.0/60.1 & 76.7/85.8 & 67.4/61.3 & 78.4/93.1 & 77.2/80.4 & 72.1/69.1 & 70.1/73.8 & 72.6/81.8 \\
    & CelebA    & 55.1/63.3 & 49.1/48.3 & 51.9/69.0 & 56.6/64.8 & 45.9/34.0 & 83.7/92.9 & 52.1/60.9 & 50.0/51.3 & 55.1/69.0 & 50.9/60.3 & 55.0/61.4\\    
    & CNNSpot   & 50.1/63.5 & 50.2/79.4 & 50.2/78.0 & 50.1/61.4 & 50.1/60.3 & 50.8/80.7 & 50.1/70.8 & 52.8/87.4 & 51.3/79.9 & 50.9/58.5 & 50.6/71.9 \\
    \midrule
    \multirow{3}{*}{DIRE}               & LSUN-B.   & 94.7/99.7 & 92.6/99.6 & 94.6/99.7 & 94.6/99.5 & 94.3/99.1 & 94.6/99.7 & 94.6/99.8 & 89.5/99.5 & 94.6/99.7 & 82.1/98.0 & 92.6/99.4 \\
    & ImageNet  & 60.2/91.3 & 54.9/86.8 & 60.3/91.7 & 57.9/89.1 & 57.6/79.6 & 58.9/90.5 & 57.5/94.0 & 40.6/64.2 & 47.6/66.1 & 28.2/61.3 & 52.3/81.5 \\
    & CelebA    & 67.8/82.7 & 62.6/62.9 & 62.4/67.1 & 75.3/97.9 & 57.4/68.0 & 74.3/93.0 & 75.2/95.0 & 67.1/93.7 & 78.3/97.0 & 54.8/39.9 & 67.5/79.7 \\
    & CNNSpot   & 74.8/86.9 & 72.3/86.3 & 65.4/81.3 & 66.1/75.2 & 52.1/56.8 & 50.1/52.1 & 55.4/58.9 & 72.9/78.3 & 53.6/65.2 & 61.3/64.9 & 62.4/70.6 \\
    
    \midrule
    \multirow{4}{*}{DNF (Ours)}                & LSUN-B.   & \textbf{100}/\textbf{100} & 99.7/\textbf{100}         & \textbf{100}/\textbf{100} & \textbf{100}/\textbf{100} & \textbf{100}/\textbf{100} & \textbf{100}/\textbf{100} & 99.8/\textbf{100}         & \textbf{100}/\textbf{100} & 99.9/\textbf{100}         & 98.9/99.9                 & 99.8/\textbf{99.9} \\ 
    & ImageNet  & \textbf{100}/\textbf{100} & \textbf{100}/\textbf{100} & \textbf{100}/\textbf{100} & 99.9/\textbf{100} & \textbf{100}/\textbf{100} & \textbf{100}/\textbf{100} & \textbf{100}/\textbf{100} & \textbf{100}/\textbf{100} & 98.8/99.2                 & \textbf{100}/\textbf{100} & \textbf{99.9}/\textbf{99.9}\\ 
    & CelebA    & \textbf{100}/\textbf{100} & 99.2/\textbf{100} & \textbf{100}/\textbf{100}         & 99.2/99.9         & \textbf{100}/\textbf{100} & \textbf{100}/\textbf{100} & \textbf{100}/\textbf{100} & 98.6/99.9 & \textbf{100}/\textbf{100} & 98.1/99.3 & 99.5/\textbf{99.9}\\ 
    & CNNSpot   & 99.7/99.7 & 97.7/99.7 & 99.9/99.7 & 99.9/99.7 & 97.7/99.7 & 82.6/99.5 & 99.9/99.7 & 90.9/99.6 & 97.5/99.7 & 99.7/99.7 & 96.5/99.7 \\ 
    \toprule
    \multirow{2}{*}{Method} & Training & \multicolumn{10}{c}{CNNSpot} & Total  \\ 
    & Dataset   & ProGAN$^\dag$   & StyleGAN  & StyleGAN2 & StarGAN3  & BigGAN    & CycleGAN  & GuaGAN    &  StarGAN  & ProjGAN   & Diff-ProjGAN & Avg.      \\
    \midrule
    \multirow{4}{*}{CNNDet.} & LSUN-B.   & 68.6/89.7 & 71.1/88.5 & 65.8/83.2 & 97.9/99.8 & 58.3/89.7 & 54.9/60.3 & 64.8/74.4 & 75.5/86.1 & 74.1/91.4 & 68.3/88.6 & 69.9/85.2 \\
    & ImageNet  & 84.4/92.8 & 82.8/88.8 & 84.3/89.2 & 50.1/61.4 & 80.3/86.4 & 57.4/53.9 & 75.3/84.1 & 94.5/98.8 & 63.3/62.2 & 59.2/56.8 & 73.2/77.4 \\
    & CelebA    & 50.3/51.6 & 54.3/62.2 & 53.7/70.4 & 97.9/99.8 & 51.1/53.2 & 50.0/48.4 & 52.3/59.3 & 50.0/44.0 & 51.8/52.8 & 52.4/55.3 & 56.4/59.7 \\    
    & CNNSpot   & \textbf{100}/\textbf{100}   & 73.4/98.5 & 68.4/97.9 & 50.1/61.4 & 59.0/88.2 & 80.7/96.8 & 79.2/98.1 & 80.9/95.4 & 52.8/90.0 & 52.0/88.3 & 69.7/91.5 \\
    \midrule
    \multirow{4}{*}{DIRE}  & LSUN-B.   & 52.8/58.8 & 51.1/56.7 & 51.7/58.0 & 84.6/99.6 & 49.7/46.9 & 49.6/50.1 & 51.3/47.4 & 47.8/40.7 & 84.6/99.6 & 84.6/99.5 & 60.8/65.7 \\
    & ImageNet  & 51.6/56.2 & 52.3/58.9 & 50.1/50.3 & 67.5/78.9 & 66.9/73.2 & 53.3/60.1 & 51.2/65.8 & 88.2/95.7 & 56.2/62.1 & 54.9/60.2 & 59.2/66.1 \\
    & CelebA    & 62.1/75.2 & 66.3/69.3 & 50.1/56.2 & 53.2/62.1 & 52.1/53.2 & 56.8/52.1 & 51.3/56.3 & 52.1/56.3 & 63.2/71.2 & 66.6/73.2 & 57.4/62.5 \\
    & CNNSpot   & 95.2/99.3 & 82.5/93.2 & 74.8/88.9 & 82.1/91.2 & 72.1/78.9 & 72.9/80.1 & 65.8/73.5 & 96.7/99.6 & 67.2/76.9 & 67.8/76.8 & 77.7/85.8 \\
    
    \midrule
    \multirow{4}{*}{DNF (Ours)}                & LSUN-B.   & 99.9/\textbf{100} & 99.3/\textbf{100} & 97.8/\textbf{100} & 99.3/\textbf{100} & \textbf{100}/\textbf{100} & \textbf{100}/\textbf{100} & \textbf{100}/\textbf{100} & \textbf{100}/\textbf{100} & 99.9/\textbf{100} & 99.9/\textbf{100} & 99.6/\textbf{100} \\ 
    & ImageNet  & \textbf{100}/\textbf{100} & 98.6/99.7 & 99.9/\textbf{100} & 99.9/\textbf{100} & \textbf{100}/\textbf{100} & \textbf{100}/\textbf{100} & 99.9/\textbf{100} & \textbf{100}/\textbf{100} & \textbf{100}/\textbf{100} & \textbf{100}/\textbf{100} & \textbf{99.8}/99.9 \\ 
    & CelebA    & \textbf{100}/\textbf{100} & \textbf{100}/\textbf{100} & 99.8/\textbf{100} & \textbf{100}/\textbf{100} & 99.9/\textbf{100} & \textbf{100}/\textbf{100} & 98.3/\textbf{100} & \textbf{100}/\textbf{100} & \textbf{100}/\textbf{100} & \textbf{100}/\textbf{100} & \textbf{99.8}/\textbf{100}\\ 
    & CNNSpot   & \textbf{100}/\textbf{100} & 99.6/\textbf{100} & 97.2/\textbf{100} & 97.7/99.7 & 90.5/\textbf{100} & 78.8/\textbf{100} & 85.5/\textbf{100} & \textbf{100}/\textbf{100} & 99.8/99.7 & 99.8/99.7 & 94.9/99.9\\ 
    
    \bottomrule
\end{tabular}}
\label{tab:gen_all}
\end{table*}

\subsection{Experimental Setup}

\noindent\textbf{Datasets.} 
To ensure a fair comparison with baselines, we conducted our experiments on three widely used datasets, DiffusionForensics~\cite{wang2023dire}, CNNSpot~\cite{wang2020cnn} and GenImage~\cite{zhu2024genimage}, which contain authentic images from sources such as ImageNet~\cite{deng2009imagenet}, LSUN-Bedroom~\cite{yu2015lsun}, and CelebA~\cite{liu2018large}, as well as images generated by various types of generative models, \eg, GANs, DMs.
These datasets are diverse and large-scale to enable comprehensive evaluation of generated image detection methods under real-world conditions. 
Concurrently, cross symbols ($\dag$) are introduced in result tables to explicitly denote generator categories included in the training set.

\vspace*{1\baselineskip} 
\noindent\textbf{Baselines.} 
We select CNNDetection~\cite{wang2020cnn}, SBI~\cite{shiohara2022detecting}, PatchForensics~\cite{chai2020makes}, F3Net~\cite{qian2020thinking}, NPR~\cite{tan2024rethinking}, LaRE$^2$~\cite{luo2024lare}, FatFormer~\cite{liu2024forgery} and DIRE~\cite{wang2023dire} as baselines, covering a range of approaches including image post-processing, frequency domain analysis, image reconstruction, and DeepFake detection. In our experiments, we make every effort to use the open-source code and models provided by their authors. We re-train these methods on our training dataset to ensure optimal performance. The re-trained approaches are denoted with an asterisk (*) in the experimental results.

\vspace*{1\baselineskip} 
\noindent\textbf{Training Details.} 
Before training, we preprocess the images following the widely adopted training settings used in CNNDetection and DIRE for training our ResNet50 classifier. First, all images are resized to a uniform size of 256×256. During training, we apply a random vertical flip to each image with a probability of 30\%. We use the Adam optimizer~\cite{kingma2014adam} with parameters $\beta_1 = 0.9$, $\beta_2 = 0.999$, a batch size of 64, and an initial learning rate of $10^{-4}$. The learning rate is reduced by a factor of 10 if the validation accuracy does not increase by 0.1\% after 5 epochs, and training is terminated when the learning rate reaches $10^{-6}$.

\vspace*{1\baselineskip} 
\noindent\textbf{Metrics.} 
We primarily evaluate our model using two metrics: \textit{Accuracy} and \textit{Average Precision (AP)}. These are two commonly used and effective metrics in generated image detection~\cite{wang2020cnn,wang2023dire}. In the tables, we report the results in terms of Accuracy / AP.

\begin{figure*}[t]
\centering
\includegraphics[width=\linewidth]{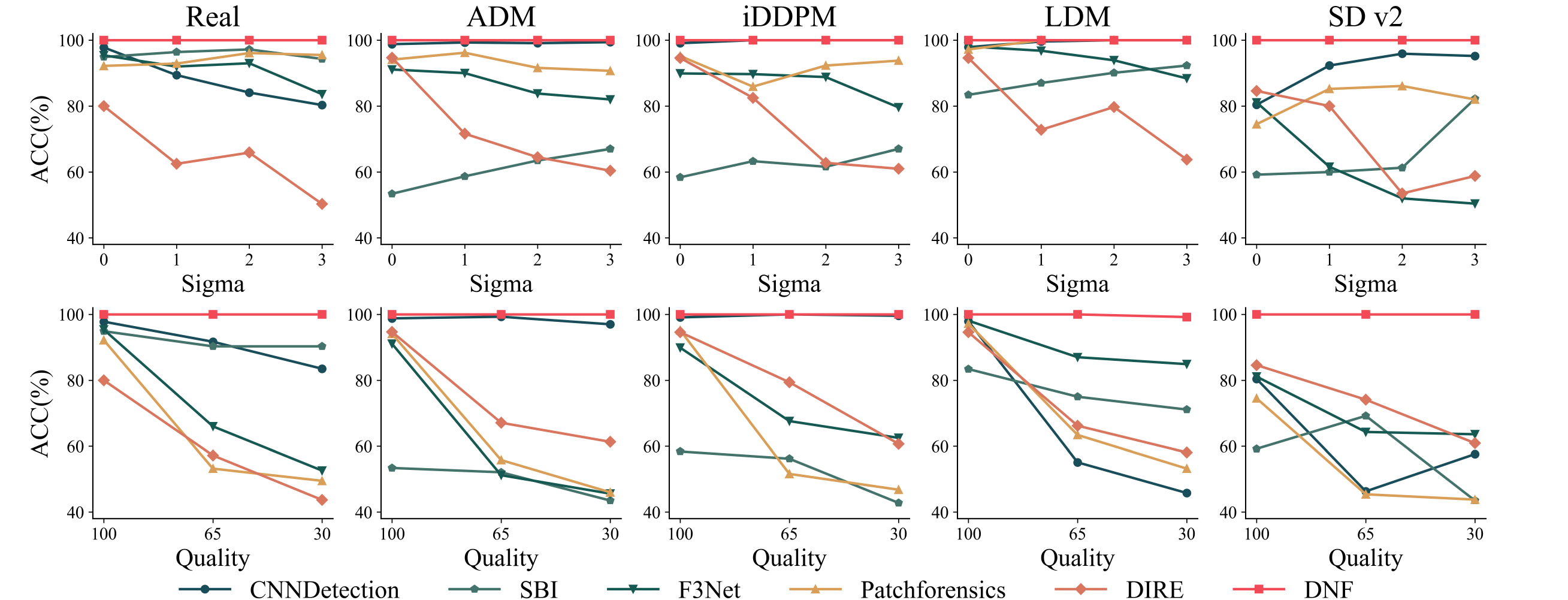}
\caption{Robustness of DNF and baseline methods to Gaussian blur and JPEG compression.}
\label{fig:robust}
\end{figure*}

\subsection{Baseline Comparison}\label{sec:baseline}

We conduct a rigorous benchmark evaluation of the DNF classifier against detection baselines on the DiffusionForensics~\cite{wang2023dire}. To ensure fair comparison, we made every effort to utilize official implementations and pretrained models for baseline methods during testing. Furthermore, we retrained CNNDetection~\cite{wang2020cnn}, Patchforensics~\cite{chai2020makes}, and F3Net~\cite{frank2020leveraging} with the same settings as the DNF classifier to demonstrate DNF classifier’s superior performance compared to these methods when using the same dataset. Comprehensive evaluation results are summarized in Table~\ref{tab:baseline}. 

Our experiments reveal some critical findings. Detection baselines including CNNDetection, PatchForensics, and SBI~\cite{shiohara2022detecting} exhibit fundamental limitations in identifying DM-generated content. While achieving high accuracy on GAN-generated images~\cite{wang2020cnn}, these methods suffer severe performance degradation when tested against generators outside the training set, with accuracy plunging to 55.8\% and AP dropping to 71.7\%.

After retraining CNNDetection, F3Net, and Patchforensics on the LSUN-Bedroom Split of DiffusionForensics, we found that these methods indeed exhibit good detection performance for images generated by DMs, achieving average accuracy of 86.0\% and average AP of 94.1\%. However, this detection performance seems to not generalize to generator unseen during training. While achieving average accuracy of 94.8\% and average AP of 99.2\% on seen generators, they only achieve average accuracy of 82.2\% and average AP of 91.9\% on generators not in training datasets. 

The most outstanding  baselines is DIRE, reaching accuracy of 92.6\% and average AP of 99.4\%, and it can generalize detection capability to the vast majority of generators. Our method, the detector trained on DNF achieves remarkably impressive performance, surpassing all previous methods on this dataset, achieving 99.8\% accuracy and 99.9\% average AP.

\subsection{Generalization Capability}\label{sec:gen}

\begin{table}[t]
\caption{
Robustness of DNF to image crop, resize and rotation.
}
\centering
\resizebox{\linewidth}{!}{
\begin{tabular}{@{\extracolsep\fill}lcccc}
    \toprule
    Perturbation & ADM & iDDPM & LDM & SD v2\\
    \midrule
    None    & \textbf{100}/\textbf{100} & \textbf{100}/\textbf{100} & \textbf{100}/\textbf{100} & \textbf{100}/\textbf{100} \\
    Crop 64 & 92.1/98.7 & 91.7/98.8 & 93.4/99.8 & 87.2/88.2\\
    Crop 224 & 99.9/\textbf{100} & \textbf{100}/\textbf{100} & \textbf{100}/\textbf{100} & \textbf{100}/\textbf{100}\\
    Resize 128 & 98.2/99.6 & 99.9/\textbf{100} & 95.2/99.9 & \textbf{100}/\textbf{100}\\
    Resize 1024 & 99.8/\textbf{100} & 99.9/\textbf{100} & \textbf{100}/\textbf{100} & 99.9/\textbf{100} \\
    Rotation $\pi/2$ & \textbf{100}/\textbf{100} & 99.9/\textbf{100} & 99.9/\textbf{100} & \textbf{100}/\textbf{100} \\
    Rotation $\pi$ & 99.9/\textbf{100} & 99.7/\textbf{100} & \textbf{100}/\textbf{100} & 99.8/\textbf{100} \\
    \bottomrule
\end{tabular}}
\label{tab:robust}
\end{table}

\begin{figure*}[t]
\centering
\includegraphics[width=\linewidth]{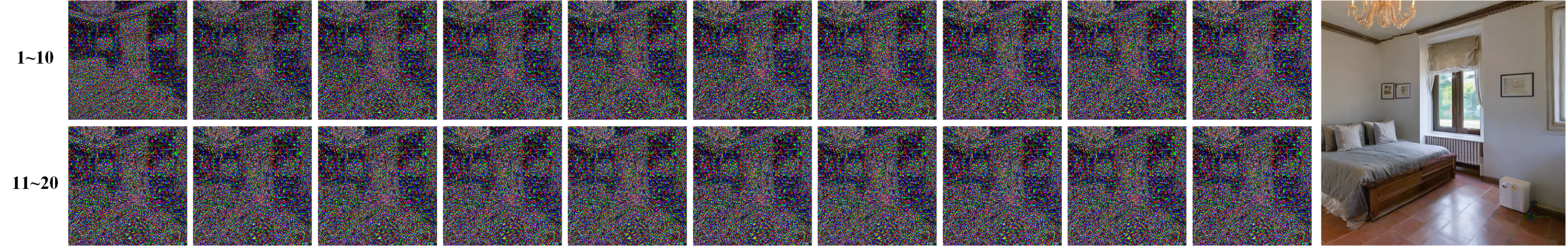}
\caption{Visualization of a sequence of estimated noises generated by a diffusion model in the inverse diffusion process.}
\label{fig:ens_vis}
\end{figure*}

\begin{table*}[t]
\caption{
Ablation studies on the impact of diffusion model $\mathcal{N}_\text{diff}$ under different model architectures $\mathcal{N}$ and pretraining datasets $\mathcal{D}_\text{pre}$.
}
\centering
\resizebox{\linewidth}{!}{
\begin{tabular}{@{}llccccccccccc@{}}
    \toprule
    Diffusion & Pretrain & \multicolumn{10}{c}{DF LSUN-Bedroom} & Total  \\ 
    Model $\mathcal{N}$ & Data $\mathcal{D}_\text{pre}$ & ADM$^\dag$ & DDPM & iDDPM$^\dag$ & LDM & PNDM$^\dag$ & SD-v2 & VQ-D & DALL-E 2 & IF & Midjourney & Avg. \\

    \midrule
    \multirow{2}{*}{DDIM}           & LSUN-B.       & \textbf{100}/\textbf{100} & 99.7/\textbf{100} & \textbf{100}/\textbf{100} & \textbf{100}/\textbf{100} & \textbf{100}/\textbf{100} & \textbf{100}/\textbf{100} & 99.8/\textbf{100} & \textbf{100}/\textbf{100} & 99.9/\textbf{100} & 98.9/99.9 & \textbf{99.8}/\textbf{99.9} \\ 
    & ImageNet      & \textbf{100}/\textbf{100} & 99.2/\textbf{100} & \textbf{100}/\textbf{100} & \textbf{100}/\textbf{100} & \textbf{100}/\textbf{100} & 99.6/99.9 & \textbf{100}/\textbf{100} & \textbf{100}/\textbf{100} & 99.6/\textbf{100} & \textbf{100}/\textbf{100} & \textbf{99.8}/\textbf{99.9}\\ 
    \multirow{2}{*}{ADM}            & LSUN-B.       & \textbf{100}/\textbf{100} & \textbf{100}/\textbf{100} & \textbf{100}/\textbf{100} & 99.8/\textbf{100} & \textbf{100}/\textbf{100} & 99.1/\textbf{100} & \textbf{100}/\textbf{100} & 98.9/99.4 & \textbf{100}/\textbf{100} & 99.2/99.9 & 99.7/\textbf{99.9} \\
    & ImageNet      & \textbf{100}/\textbf{100} & \textbf{100}/\textbf{100} & \textbf{100}/\textbf{100} & 97.2/98.9 & \textbf{100}/\textbf{100} & 99.2/99.9 & \textbf{100}/\textbf{100} & 99.8/\textbf{100} & 99.7/\textbf{100} & \textbf{100}/\textbf{100} & 99.5/99.8 \\ 
    \bottomrule
\end{tabular}}
\label{tab:ablation_dm}
\end{table*}
\begin{table}[t]
\label{tab：ablation_ffs}
\caption{Ablation studies on impact of feature fusion $\boldsymbol{g}$.} 
\resizebox{\linewidth}{!}{
\begin{tabular}{@{\extracolsep\fill}lcccccc}
    \toprule
    Method    & ADM & iDDPM & LDM & StyleGAN & SD-v1 & Avg.\\
    \midrule
    \texttt{FIRST}   & \textbf{100}/\textbf{100} & \textbf{100}/\textbf{100} & \textbf{100}/\textbf{100} & 99.3/\textbf{100} & 96.3/\textbf{100} & 99.1/\textbf{100}     \\
    \texttt{MEAN}     & 98.2/99.9                 & 99.1/\textbf{100}         & 97.5/\textbf{100}         & 98.2/99.9         & 98.2/\textbf{100} & 98.2/99.9\\
    \texttt{CONV}    & \textbf{100}/\textbf{100} & \textbf{100}/\textbf{100} & 99.7/99.9 & \textbf{99.6}/99.9         & \textbf{100}/\textbf{100} & \textbf{99.8}/99.9\\ 
    \bottomrule
\end{tabular}}
\end{table}

In generalization capability evaluation, we selected the best-performing CNNDetection~\cite{wang2020cnn} after retraining and the overall best-performing DIRE~\cite{wang2023dire} to conduct a generalization evaluation experiment with our DNF. In this experiment, each method will be retrained on three training splits of DiffusionForensics~\cite{wang2023dire} and CNNSpot~\cite{wang2020cnn} and tested on five test set from DiffusionForensics, CNNSpot and GenImage~\cite{zhu2024genimage} to assess the methods' cross-dataset and cross-generator generalization capabilities. 
The evaluation results across multiple datasets can be found in the Table~\ref{tab:gen_all}. 
To address overlapping generator categories (\eg, Midjourney, ADM, VQDM, BigGAN) between GenImage and other datasets, we systematically consolidate redundant results into non-GenImage benchmark splits for streamlined presentation.

\subsubsection{Cross-dataset Generalization} 

When the training and testing datasets contain similar image content, all three methods perform excellently. Taking DIRE as an example, the DIRE detector trained on the LSUN-Bedroom Split or CelebA Split achieves accuracies of 92.6\% and 95.4\%, respectively, on the corresponding test sets. However, when the test set contains different content, the performance of the baseline methods drops significantly. Specifically, the accuracy of DIRE in cross-validation between LSUN-Bedroom Split and CelebA Split drops to 50.2\% and 67.5\%, respectively. In contrast, DNF exhibits excellent cross-dataset generalization by capturing the universal gap between real and generated images. It achieves an average accuracy of 96.2\% and an average AP of 99.8\% in cross-dataset validation.

\subsubsection{Cross-generator Generalization} 

Another important metric for evaluating generated image detection is the ability to detect images generated by unknown generators not included in the training set. CNNDetection trained on CNNSpot can detect images generated by ProGAN~\cite{karras2017progressive} with 100\% accuracy, but when detecting other generators such as StyleGAN~\cite{karras2019style}, the accuracy drops to 66.3\%. DIRE trained on the ImageNet Split achieves a 99.8\% accuracy in detecting images generated by ADM~\cite{dhariwal2021diffusion}, and 96.6\% accuracy in Stable Diffusions, but when detecting images from other generators, \eg, Glide, wukong, it only achieves an 60.0\% accuracy. Meanwhile, DNF achieves detection accuracies for unseen generators of 96.2\%, 99.8\%, 99.6\%, and 78.7\% across these three test sets. Crucially, according to the comprehensive result in supplementary material, DNF can even generalize between generators with different principles.

\subsection{Perturbation Robustness}\label{sec:robust}

When images are shared on social networks, they often undergo perturbations such as Gaussian blur or JPEG compression, resulting in the loss of image details. This loss significantly impacts the performance of generated image detection. We designed robustness experiments to evaluate the robustness of various methods under such perturbations.
We applied varying degrees of perturbation to the images, including Gaussian blur with $\sigma$ $\in$ \{0, 1, 2, 3\} and JPEG compression with $Quality$ $\in$ \{100, 65, 30\}, to explore the performance fluctuations of different methods as perturbation intensity increases. In Figure~\ref{fig:robust}, as the perturbation intensifies, the detailed information in the images becomes less distinct, making detection more challenging. DNF, with its inherent ability to enhance image details, ensures that even weakened details remain effective for image generation detection. As a result, DNF showed minimal performance degradation, consistently achieving detection accuracy above 99.2\%.

Additionally, we assessed the robustness of our method against disturbances such as resizing, cropping, and rotation to evaluate their impact on DNF's performance. We present the results in Table~\ref{tab:robust}. Since these transformations are common both in image transmission and data augmentation, they did not significantly degrade most methods' performance. However, excessive cropping of image content caused a noticeable performance drop in DNF.

\subsection{Ablation Studies}\label{sec:ablation}

\noindent\textbf{Impact of DM.} 
Algorithm~\ref{alg:obtain_ens} requires a pre-trained DM $\mathcal{N}_\text{diff}$ to generate estimate noise. Different DM structures~\cite{song2020denoising,dhariwal2021diffusion} and pre-training datasets~\cite{liu2018large,deng2009imagenet} may affect the final DNF. To investigate this impact, we design experiments, and the results are presented in Table~\ref{tab:ablation_dm}.
We observed that DM $\mathcal{N}_\text{diff}$ with different structures $\mathcal{N}$ or pre-training on different datasets $\mathcal{D}_\text{pre}$ does not significantly impact the DNF. However, based on the design principles of DNF and issues observed in the experiments, we emphasize that $\mathcal{N}_\text{diff}$ should be pre-trained until it sufficiently converges to capture high-frequency details in the images. Inadequate pre-training of the $\mathcal{N}_\text{diff}$ results in a homogeneous pure noise output, which fails to reflect the gap between different data distributions.

\begin{table}[t]
\caption{Ablation study on the impact of different image storage formats.}
\label{tab8}
\begin{tabular*}{\columnwidth}{@{\extracolsep\fill}lccc}
    \toprule
    & Original & PNG & JPEG \\
    \midrule
    Original    & \textbf{98.6}/\textbf{99.9} & 93.1/98.2 & 95.7/99.7 \\
    PNG         & 96.4/\textbf{99.9} & \textbf{98.9}/\textbf{99.9} & 94.3/99.7 \\
    JPEG        & 94.3/99.6 & 86.9/92.5 & \textbf{96.1}/\textbf{99.9} \\
    \bottomrule
\end{tabular*}
\end{table}

\vspace*{1\baselineskip} 
\noindent\textbf{Impact of Fusion Strategy.} 
How the estimated noise sequence $\{\epsilon_i\}$ is processed fundamentally impacts the performance of the DNF. In this study, we investigate the effect of different feature fusion strategies $\boldsymbol{g}$ on DNF performance. The experimental results are presented in Table~\ref{tab：ablation_ffs}. Notably, the \texttt{MEAN} strategy shows a slight performance drop compared to \texttt{FIRST}. In Figure~\ref{fig:ens_vis}, we visualize an estimated noise sequence, where later estimates are closer to pure noise. The simple averaging in the \texttt{MEAN} strategy may introduce unnecessary interference. The \texttt{CONV} strategy achieves the best performance, as convolution better captures the varying features throughout the sequence. Additionally, although \texttt{FIRST} performs slightly worse, it offers superior overall inference speed due to requiring fewer computations to estimate the noise.

\vspace*{1\baselineskip} 
\noindent\textbf{Impact of Image Format.} 
Prior research~\cite{grommelt2024fake} has demonstrated that image storage formats may introduce unintended biases to classifiers.
In our format‐ablation experiments, we trained classifiers on LSUN‑Bedroom splits saved as either JPEG or PNG and evaluated them on ImageNet and CelebA. We observed that matching train/test formats improved accuracy, confirming that image formats introduce bias. To mitigate this, we standardized training data by converting all originals to PNG before DNF generation, while leaving test images in their native formats. Despite format variations, our classifier maintained strong performance, indicating minimal format-induced bias. The preprocessing strategy effectively decouples data format from authenticity, ensuring fair evaluation.

\section{Conclusion}\label{sec:conclusion}

In this paper, for the first time, we leverage estimated noises in the inverse diffusion process to design a novel image representation, DNF, specifically for detecting generated images. The classifiers trained on DNF exhibit significantly improved detection performance, enhanced generalization, and greater robustness to perturbations, surpassing baseline methods. Our approach provides a more effective solution for identifying generated images. We believe our method can serve as a valuable contribution to addressing the emerging risks associated with generated images, enabling more proactive and efficient safeguards in practical applications.\


\newpage

\begin{ack}
This work is supported by the Natural Science Foundation of Zhejiang Pvovince, China under No. LD24F020002.
\end{ack}


\bibliography{bibfile}

\begin{thebibliography}{55}
\providecommand{\natexlab}[1]{#1}
\providecommand{\url}[1]{\texttt{#1}}
\expandafter\ifx\csname urlstyle\endcsname\relax
  \providecommand{\doi}[1]{doi: #1}\else
  \providecommand{\doi}{doi: \begingroup \urlstyle{rm}\Url}\fi

\bibitem[Betker et~al.(2023)Betker, Goh, Jing, Brooks, Wang, Li, Ouyang, Zhuang, Lee, Guo, et~al.]{betker2023improving}
J.~Betker, G.~Goh, L.~Jing, T.~Brooks, J.~Wang, L.~Li, L.~Ouyang, J.~Zhuang, J.~Lee, Y.~Guo, et~al.
\newblock Improving image generation with better captions.
\newblock \emph{Computer Science. https://cdn. openai. com/papers/dall-e-3. pdf}, 2\penalty0 (3):\penalty0 8, 2023.

\bibitem[Brock(2018)]{brock2018large}
A.~Brock.
\newblock Large scale gan training for high fidelity natural image synthesis.
\newblock \emph{arXiv preprint arXiv:1809.11096}, 2018.

\bibitem[Cao et~al.(2022)Cao, Ma, Yao, Chen, Ding, and Yang]{cao2022end}
J.~Cao, C.~Ma, T.~Yao, S.~Chen, S.~Ding, and X.~Yang.
\newblock End-to-end reconstruction-classification learning for face forgery detection.
\newblock In \emph{Proceedings of the IEEE/CVF Conference on Computer Vision and Pattern Recognition}, pages 4113--4122, 2022.

\bibitem[Chai et~al.(2020)Chai, Bau, Lim, and Isola]{chai2020makes}
L.~Chai, D.~Bau, S.-N. Lim, and P.~Isola.
\newblock What makes fake images detectable? understanding properties that generalize.
\newblock In \emph{Computer Vision--ECCV 2020: 16th European Conference, Glasgow, UK, August 23--28, 2020, Proceedings, Part XXVI 16}, pages 103--120. Springer, 2020.

\bibitem[Chen et~al.(2024)Chen, Zeng, Yang, and Yang]{chendrct}
B.~Chen, J.~Zeng, J.~Yang, and R.~Yang.
\newblock Drct: Diffusion reconstruction contrastive training towards universal detection of diffusion generated images.
\newblock In \emph{Forty-first International Conference on Machine Learning}, 2024.

\bibitem[Choi et~al.(2018)Choi, Choi, Kim, Ha, Kim, and Choo]{choi2018stargan}
Y.~Choi, M.~Choi, M.~Kim, J.-W. Ha, S.~Kim, and J.~Choo.
\newblock Stargan: Unified generative adversarial networks for multi-domain image-to-image translation.
\newblock In \emph{Proceedings of the IEEE conference on computer vision and pattern recognition}, pages 8789--8797, 2018.

\bibitem[Corvi et~al.(2023)Corvi, Cozzolino, Zingarini, Poggi, Nagano, and Verdoliva]{corvi2023detection}
R.~Corvi, D.~Cozzolino, G.~Zingarini, G.~Poggi, K.~Nagano, and L.~Verdoliva.
\newblock On the detection of synthetic images generated by diffusion models.
\newblock In \emph{ICASSP 2023-2023 IEEE International Conference on Acoustics, Speech and Signal Processing (ICASSP)}. IEEE, 2023.

\bibitem[Deng et~al.(2009)Deng, Dong, Socher, Li, Li, and Fei-Fei]{deng2009imagenet}
J.~Deng, W.~Dong, R.~Socher, L.-J. Li, K.~Li, and L.~Fei-Fei.
\newblock Imagenet: A large-scale hierarchical image database.
\newblock In \emph{2009 IEEE conference on computer vision and pattern recognition}, pages 248--255. Ieee, 2009.

\bibitem[Dhariwal and Nichol(2021)]{dhariwal2021diffusion}
P.~Dhariwal and A.~Nichol.
\newblock Diffusion models beat gans on image synthesis.
\newblock \emph{Advances in neural information processing systems}, 34:\penalty0 8780--8794, 2021.

\bibitem[Frank et~al.(2020)Frank, Eisenhofer, Sch{\"o}nherr, Fischer, Kolossa, and Holz]{frank2020leveraging}
J.~Frank, T.~Eisenhofer, L.~Sch{\"o}nherr, A.~Fischer, D.~Kolossa, and T.~Holz.
\newblock Leveraging frequency analysis for deep fake image recognition.
\newblock In \emph{International conference on machine learning}, pages 3247--3258. PMLR, 2020.

\bibitem[Giglietto et~al.(2019)Giglietto, Iannelli, Valeriani, and Rossi]{giglietto2019fake}
F.~Giglietto, L.~Iannelli, A.~Valeriani, and L.~Rossi.
\newblock ‘fake news’ is the invention of a liar: How false information circulates within the hybrid news system.
\newblock \emph{Current sociology}, 67\penalty0 (4):\penalty0 625--642, 2019.

\bibitem[Goodfellow et~al.(2020)Goodfellow, Pouget-Abadie, Mirza, Xu, Warde-Farley, Ozair, Courville, and Bengio]{goodfellow2020generative}
I.~Goodfellow, J.~Pouget-Abadie, M.~Mirza, B.~Xu, D.~Warde-Farley, S.~Ozair, A.~Courville, and Y.~Bengio.
\newblock Generative adversarial networks.
\newblock \emph{Communications of the ACM}, 63\penalty0 (11):\penalty0 139--144, 2020.

\bibitem[Grommelt et~al.(2024)Grommelt, Weiss, Pfreundt, and Keuper]{grommelt2024fake}
P.~Grommelt, L.~Weiss, F.-J. Pfreundt, and J.~Keuper.
\newblock Fake or jpeg? revealing common biases in generated image detection datasets.
\newblock \emph{arXiv preprint arXiv:2403.17608}, 2024.

\bibitem[Gu et~al.(2022)Gu, Chen, Bao, Wen, Zhang, Chen, Yuan, and Guo]{gu2022vector}
S.~Gu, D.~Chen, J.~Bao, F.~Wen, B.~Zhang, D.~Chen, L.~Yuan, and B.~Guo.
\newblock Vector quantized diffusion model for text-to-image synthesis.
\newblock In \emph{Proceedings of the IEEE/CVF conference on computer vision and pattern recognition}, pages 10696--10706, 2022.

\bibitem[Ho et~al.(2020)Ho, Jain, and Abbeel]{ho2020denoising}
J.~Ho, A.~Jain, and P.~Abbeel.
\newblock Denoising diffusion probabilistic models.
\newblock \emph{Advances in neural information processing systems}, 2020.

\bibitem[Karras(2017)]{karras2017progressive}
T.~Karras.
\newblock Progressive growing of gans for improved quality, stability, and variation.
\newblock \emph{arXiv preprint arXiv:1710.10196}, 2017.

\bibitem[Karras et~al.(2019)Karras, Laine, and Aila]{karras2019style}
T.~Karras, S.~Laine, and T.~Aila.
\newblock A style-based generator architecture for generative adversarial networks.
\newblock In \emph{Proceedings of the IEEE/CVF conference on computer vision and pattern recognition}, 2019.

\bibitem[Karras et~al.(2020)Karras, Laine, Aittala, Hellsten, Lehtinen, and Aila]{karras2020analyzing}
T.~Karras, S.~Laine, M.~Aittala, J.~Hellsten, J.~Lehtinen, and T.~Aila.
\newblock Analyzing and improving the image quality of stylegan.
\newblock In \emph{Proceedings of the IEEE/CVF conference on computer vision and pattern recognition}, pages 8110--8119, 2020.

\bibitem[Karras et~al.(2021)Karras, Aittala, Laine, H{\"a}rk{\"o}nen, Hellsten, Lehtinen, and Aila]{karras2021alias}
T.~Karras, M.~Aittala, S.~Laine, E.~H{\"a}rk{\"o}nen, J.~Hellsten, J.~Lehtinen, and T.~Aila.
\newblock Alias-free generative adversarial networks.
\newblock \emph{Advances in neural information processing systems}, 34:\penalty0 852--863, 2021.

\bibitem[Kingma(2013)]{kingma2013auto}
D.~P. Kingma.
\newblock Auto-encoding variational bayes.
\newblock \emph{arXiv preprint arXiv:1312.6114}, 2013.

\bibitem[Kingma and Ba(2015)]{kingma2014adam}
D.~P. Kingma and J.~Ba.
\newblock Adam: A method for stochastic optimization.
\newblock \emph{International Conference on Learning Representations}, 2015.

\bibitem[Li et~al.(2023)Li, Liu, Lian, Yang, Dong, Kang, Zhang, and Keutzer]{li2023q}
X.~Li, Y.~Liu, L.~Lian, H.~Yang, Z.~Dong, D.~Kang, S.~Zhang, and K.~Keutzer.
\newblock Q-diffusion: Quantizing diffusion models.
\newblock In \emph{Proceedings of the IEEE/CVF International Conference on Computer Vision}, pages 17535--17545, 2023.

\bibitem[Liu et~al.(2024)Liu, Tan, Tan, Wei, Wang, and Zhao]{liu2024forgery}
H.~Liu, Z.~Tan, C.~Tan, Y.~Wei, J.~Wang, and Y.~Zhao.
\newblock Forgery-aware adaptive transformer for generalizable synthetic image detection.
\newblock In \emph{Proceedings of the IEEE/CVF Conference on Computer Vision and Pattern Recognition}, pages 10770--10780, 2024.

\bibitem[Liu et~al.(2022)Liu, Ren, Lin, and Zhao]{liu2022pseudo}
L.~Liu, Y.~Ren, Z.~Lin, and Z.~Zhao.
\newblock Pseudo numerical methods for diffusion models on manifolds.
\newblock \emph{arXiv preprint arXiv:2202.09778}, 2022.

\bibitem[Liu et~al.(2018)Liu, Luo, Wang, and Tang]{liu2018large}
Z.~Liu, P.~Luo, X.~Wang, and X.~Tang.
\newblock Large-scale celebfaces attributes (celeba) dataset.
\newblock \emph{Retrieved August}, 15\penalty0 (2018):\penalty0 11, 2018.

\bibitem[Luo et~al.(2024)Luo, Du, Yan, and Ding]{luo2024lare}
Y.~Luo, J.~Du, K.~Yan, and S.~Ding.
\newblock Lare\^{} 2: Latent reconstruction error based method for diffusion-generated image detection.
\newblock In \emph{Proceedings of the IEEE/CVF Conference on Computer Vision and Pattern Recognition}, pages 17006--17015, 2024.

\bibitem[{Midjourney}(2022)]{midjourney2022}
{Midjourney}.
\newblock {Midjourney}.
\newblock {\url{https://www.midjourney.com}}, 2022.

\bibitem[Murdoch(2021)]{murdoch2021privacy}
B.~Murdoch.
\newblock Privacy and artificial intelligence: challenges for protecting health information in a new era.
\newblock \emph{BMC Medical Ethics}, 22:\penalty0 1--5, 2021.

\bibitem[Nichol et~al.(2021)Nichol, Dhariwal, Ramesh, Shyam, Mishkin, McGrew, Sutskever, and Chen]{nichol2021glide}
A.~Nichol, P.~Dhariwal, A.~Ramesh, P.~Shyam, P.~Mishkin, B.~McGrew, I.~Sutskever, and M.~Chen.
\newblock Glide: Towards photorealistic image generation and editing with text-guided diffusion models.
\newblock \emph{arXiv preprint arXiv:2112.10741}, 2021.

\bibitem[Nichol and Dhariwal(2021)]{nichol2021improved}
A.~Q. Nichol and P.~Dhariwal.
\newblock Improved denoising diffusion probabilistic models.
\newblock In \emph{International conference on machine learning}, pages 8162--8171. PMLR, 2021.

\bibitem[Park et~al.(2019)Park, Liu, Wang, and Zhu]{park2019semantic}
T.~Park, M.-Y. Liu, T.-C. Wang, and J.-Y. Zhu.
\newblock Semantic image synthesis with spatially-adaptive normalization.
\newblock In \emph{Proceedings of the IEEE/CVF conference on computer vision and pattern recognition}, pages 2337--2346, 2019.

\bibitem[Peebles and Xie(2023)]{peebles2023scalable}
W.~Peebles and S.~Xie.
\newblock Scalable diffusion models with transformers.
\newblock In \emph{Proceedings of the IEEE/CVF International Conference on Computer Vision}, pages 4195--4205, 2023.

\bibitem[Phung et~al.(2023)Phung, Dao, and Tran]{phung2023wavelet}
H.~Phung, Q.~Dao, and A.~Tran.
\newblock Wavelet diffusion models are fast and scalable image generators.
\newblock In \emph{Proceedings of the IEEE/CVF conference on computer vision and pattern recognition}, pages 10199--10208, 2023.

\bibitem[Qi et~al.(2019)Qi, Cao, Yang, Guo, and Li]{qi2019exploiting}
P.~Qi, J.~Cao, T.~Yang, J.~Guo, and J.~Li.
\newblock Exploiting multi-domain visual information for fake news detection.
\newblock In \emph{2019 IEEE international conference on data mining (ICDM)}, pages 518--527. IEEE, 2019.

\bibitem[Qian et~al.(2020)Qian, Yin, Sheng, Chen, and Shao]{qian2020thinking}
Y.~Qian, G.~Yin, L.~Sheng, Z.~Chen, and J.~Shao.
\newblock Thinking in frequency: Face forgery detection by mining frequency-aware clues.
\newblock In \emph{European conference on computer vision}, pages 86--103. Springer, 2020.

\bibitem[Ramesh et~al.(2022)Ramesh, Dhariwal, Nichol, Chu, and Chen]{ramesh2022hierarchical}
A.~Ramesh, P.~Dhariwal, A.~Nichol, C.~Chu, and M.~Chen.
\newblock Hierarchical text-conditional image generation with clip latents.
\newblock \emph{arXiv preprint arXiv:2204.06125}, 1\penalty0 (2):\penalty0 3, 2022.

\bibitem[Rana et~al.(2022)Rana, Nobi, Murali, and Sung]{rana2022deepfake}
M.~S. Rana, M.~N. Nobi, B.~Murali, and A.~H. Sung.
\newblock Deepfake detection: A systematic literature review.
\newblock \emph{IEEE access}, 10:\penalty0 25494--25513, 2022.

\bibitem[Rombach et~al.(2022)Rombach, Blattmann, Lorenz, Esser, and Ommer]{rombach2022high}
R.~Rombach, A.~Blattmann, D.~Lorenz, P.~Esser, and B.~Ommer.
\newblock High-resolution image synthesis with latent diffusion models.
\newblock In \emph{Proceedings of the IEEE/CVF conference on computer vision and pattern recognition}, pages 10684--10695, 2022.

\bibitem[Saharia et~al.(2022)Saharia, Chan, Saxena, Li, Whang, Denton, Ghasemipour, Gontijo~Lopes, Karagol~Ayan, Salimans, et~al.]{saharia2022photorealistic}
C.~Saharia, W.~Chan, S.~Saxena, L.~Li, J.~Whang, E.~L. Denton, K.~Ghasemipour, R.~Gontijo~Lopes, B.~Karagol~Ayan, T.~Salimans, et~al.
\newblock Photorealistic text-to-image diffusion models with deep language understanding.
\newblock \emph{Advances in neural information processing systems}, 35:\penalty0 36479--36494, 2022.

\bibitem[Shi et~al.(2023)Shi, Chen, Chen, and Zhang]{shi2023discrepancy}
Z.~Shi, H.~Chen, L.~Chen, and D.~Zhang.
\newblock Discrepancy-guided reconstruction learning for image forgery detection.
\newblock \emph{arXiv preprint arXiv:2304.13349}, 2023.

\bibitem[Shiohara and Yamasaki(2022)]{shiohara2022detecting}
K.~Shiohara and T.~Yamasaki.
\newblock Detecting deepfakes with self-blended images.
\newblock In \emph{Proceedings of the IEEE/CVF Conference on Computer Vision and Pattern Recognition}, pages 18720--18729, 2022.

\bibitem[Sinitsa and Fried(2023)]{sinitsa2023deep}
S.~Sinitsa and O.~Fried.
\newblock Deep image fingerprint: Accurate and low budget synthetic image detector.
\newblock \emph{arXiv preprint arXiv:2303.10762}, 2023.

\bibitem[Song et~al.(2020)Song, Meng, and Ermon]{song2020denoising}
J.~Song, C.~Meng, and S.~Ermon.
\newblock Denoising diffusion implicit models.
\newblock \emph{arXiv preprint arXiv:2010.02502}, 2020.

\bibitem[Tan et~al.(2023)Tan, Zhao, Wei, Gu, and Wei]{tan2023learning}
C.~Tan, Y.~Zhao, S.~Wei, G.~Gu, and Y.~Wei.
\newblock Learning on gradients: Generalized artifacts representation for gan-generated images detection.
\newblock In \emph{Proceedings of the IEEE/CVF Conference on Computer Vision and Pattern Recognition}, pages 12105--12114, 2023.

\bibitem[Tan et~al.(2024)Tan, Zhao, Wei, Gu, Liu, and Wei]{tan2024rethinking}
C.~Tan, Y.~Zhao, S.~Wei, G.~Gu, P.~Liu, and Y.~Wei.
\newblock Rethinking the up-sampling operations in cnn-based generative network for generalizable deepfake detection.
\newblock In \emph{Proceedings of the IEEE/CVF Conference on Computer Vision and Pattern Recognition}, pages 28130--28139, 2024.

\bibitem[Uyyala and Yadav(2023)]{uyyala2023advanced}
P.~Uyyala and D.~C. Yadav.
\newblock The advanced proprietary ai/ml solution as anti-fraudtensorlink4cheque (aftl4c) for cheque fraud detection.
\newblock \emph{The International journal of analytical and experimental modal analysis}, 15\penalty0 (4):\penalty0 1914--1921, 2023.

\bibitem[Van~der Maaten and Hinton(2008)]{van2008visualizing}
L.~Van~der Maaten and G.~Hinton.
\newblock Visualizing data using t-sne.
\newblock \emph{Journal of machine learning research}, 9\penalty0 (11), 2008.

\bibitem[Wang et~al.(2020)Wang, Wang, Zhang, Owens, and Efros]{wang2020cnn}
S.-Y. Wang, O.~Wang, R.~Zhang, A.~Owens, and A.~A. Efros.
\newblock Cnn-generated images are surprisingly easy to spot... for now.
\newblock In \emph{Proceedings of the IEEE/CVF conference on computer vision and pattern recognition}, pages 8695--8704, 2020.

\bibitem[Wang et~al.(2023)Wang, Bao, Zhou, Wang, Hu, Chen, and Li]{wang2023dire}
Z.~Wang, J.~Bao, W.~Zhou, W.~Wang, H.~Hu, H.~Chen, and H.~Li.
\newblock Dire for diffusion-generated image detection.
\newblock In \emph{Proceedings of the IEEE/CVF International Conference on Computer Vision}, 2023.

\bibitem[Westerlund(2019)]{westerlund2019emergence}
M.~Westerlund.
\newblock The emergence of deepfake technology: A review.
\newblock \emph{Technology innovation management review}, 9\penalty0 (11), 2019.

\bibitem[{wukong}(2023)]{wukong2023}
{wukong}.
\newblock {wukong}.
\newblock {\url{https://xihe.mindspore.cn/modelzoo/wukong}}, 2023.

\bibitem[Yu et~al.(2015)Yu, Seff, Zhang, Song, Funkhouser, and Xiao]{yu2015lsun}
F.~Yu, A.~Seff, Y.~Zhang, S.~Song, T.~Funkhouser, and J.~Xiao.
\newblock Lsun: Construction of a large-scale image dataset using deep learning with humans in the loop.
\newblock \emph{arXiv preprint arXiv:1506.03365}, 2015.

\bibitem[Zhong et~al.(2023)Zhong, Xu, Qian, and Zhang]{zhong2023rich}
N.~Zhong, Y.~Xu, Z.~Qian, and X.~Zhang.
\newblock Rich and poor texture contrast: A simple yet effective approach for ai-generated image detection.
\newblock \emph{arXiv preprint arXiv:2311.12397}, 2023.

\bibitem[Zhu et~al.(2017)Zhu, Park, Isola, and Efros]{zhu2017unpaired}
J.-Y. Zhu, T.~Park, P.~Isola, and A.~A. Efros.
\newblock Unpaired image-to-image translation using cycle-consistent adversarial networks.
\newblock In \emph{Proceedings of the IEEE international conference on computer vision}, pages 2223--2232, 2017.

\bibitem[Zhu et~al.(2024)Zhu, Chen, Yan, Huang, Lin, Li, Tu, Hu, Hu, and Wang]{zhu2024genimage}
M.~Zhu, H.~Chen, Q.~Yan, X.~Huang, G.~Lin, W.~Li, Z.~Tu, H.~Hu, J.~Hu, and Y.~Wang.
\newblock Genimage: A million-scale benchmark for detecting ai-generated image.
\newblock \emph{Advances in Neural Information Processing Systems}, 36, 2024.

\end{thebibliography}


\appendix
\section{Discussion}

\subsection{Frequency Analysis of DNF}

\begin{figure*}[t]
\centering
\includegraphics[width=\linewidth]{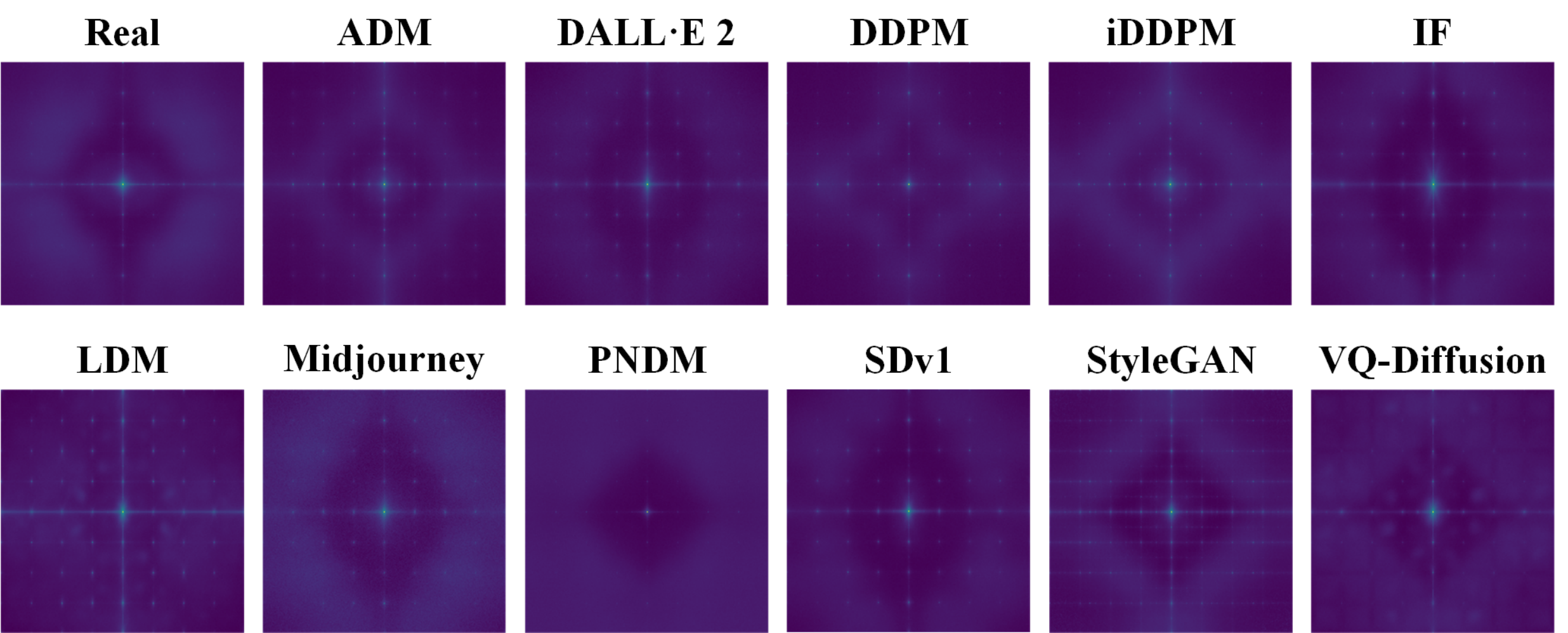}
\caption{Frequency-domain visualization of DNF for images generated by different models.}
\label{fig:freq_vis}
\end{figure*}

From a frequency-domain perspective, DNF amplifies the gap between real and generated images. We demonstrate this by visualizing the frequency-domain features of images from different sources. In the frequency-domain analysis, we randomly selected 1,000 images generated by each type of generator, all depicting the "Bedroom" scene, and applied Fourier Transform to them. We then computed the mean of the transformed images and visualized the results in Figure~\ref{fig:freq_vis}. This approach allows us to capture and compare the global spectral patterns introduced by different generative models.

Our findings reveal significant differences in the frequency domains of DNF across different data sources, with a clear distinction between real and generated images. The frequency components of real images are more evenly distributed and natural, while those of generated images exhibit periodic artifacts. This result provides an additional explanation for the strong performance of DNF, highlighting its ability to distinguish between real and generated images based on their frequency-domain characteristics. These observations confirm that frequency-domain cues, though often overlooked, can serve as reliable signals for detecting generative artifacts. 

\subsection{High-Dimensional Representation Visualization}

We used t-SNE~\cite{van2008visualizing} to visualize the high-dimensional representations learned by the classifier, thereby demonstrating the superiority of DNF as a classification feature. As shown in Table~\ref{fig:tsne_vis}, in the binary classification task between real and generated images, the high-dimensional latent representations of real and generated samples were clearly separated. This indicates that DNF effectively captures distinguishing features between these two classes. In the multi-class data source classification task, we observed that samples from different data sources were also clearly distinguishable, which can be attributed to the unique features exhibited by DNF across various data sources. This result further highlights the capability of DNF to differentiate images from diverse origins. 

In contrast, features generated by other methods~\cite{wang2020cnn,wang2023dire} for classification were only able to roughly separate real and fake samples, failing to achieve the same level of distinction. This visualization provides strong evidence of the effectiveness of DNF in distinguishing between different types of images and data sources.

\begin{figure*}[t]
    \centering
    \includegraphics[width=\linewidth]{assets/figure_5.pdf}
    \caption{t-SNE visualization of DNF, DIRE, CNNDetection features, and original images.}
    \label{fig:tsne_vis}
\end{figure*}

\section{Experimental Setup Details}
\subsection{Dataset}

We will introduce the three datasets used in our experiments, \textbf{CNNSpot}~\cite{wang2020cnn}, \textbf{DiffusionForensics}~\cite{wang2023dire} and \textbf{GenImage}~\cite{zhu2024genimage}. These are widely recognized datasets in generated image detection.

\textbf{CNNSpot} is a comprehensive dataset that features a multitude of images generated by various CNN-based architectures, predominantly GANs. It encompasses three leading unconditional GANs: ProGAN~\cite{karras2017progressive}, StyleGAN~\cite{karras2019style,karras2020analyzing,karras2021alias}, and BigGAN~\cite{brock2018large}, each trained on either the LSUN or ImageNet datasets. These models exhibit distinct network structures and training methodologies. ProGAN and StyleGAN are designed to handle different categories, with StyleGAN incorporating substantial per-pixel noise to enhance high-frequency details. BigGAN, on the other hand, adopts a unified, class-conditional framework, is optimized with very large batch sizes, and incorporates self-attention mechanisms. Additionally, CNNSpot includes three conditional GANs: GauGAN~\cite{park2019semantic}, a state-of-the-art image-to-image translation tool; CycleGAN~\cite{zhu2017unpaired}, a popular method for unpaired image translation; and StarGAN~\cite{choi2018stargan}, another well-regarded approach for similar tasks.

\begin{table}[t]
    \caption{The dataset composition of CNNSpot.}
    \centering
    \resizebox{1.0\linewidth}{!}{
    \begin{tabular}{lllr}
        \toprule
        Category & Generator & Image Source & \# Images \\
        \midrule
        \multirow{3}{*}{Unconditional GAN}  & ProGAN & LSUN & 8.0k \\
                                            & StyleGAN & LSUN & 12.0k \\
                                            & BigGAN & ImageNet & 8.0k \\ 
        \midrule
        \multirow{3}{*}{Conditional GAN}  & CycleGAN & Style/Object tansfer & 2.6k \\
                                            & StarGAN & CelebA & 4.0k \\
                                            & GauGAN & COCO & 10.0k \\ 
        \bottomrule
    \end{tabular}} 
\label{tab:dataset_cnn}
\end{table}
\begin{table}[t]
\caption{The dataset composition of DiffusionForensics.}
\centering
\resizebox{1.0\linewidth}{!}{
\begin{tabular}{lllr}
    \toprule
    Image Source & Category & Generator & \# Images \\
    \midrule
    \multirow{12}{*}{LSUN-Bedroom}  & Real & None & 4.2k \\
    \cmidrule{2-4}
        & \multirow{3}{*}{Unconditional} & ADM & 4.2k \\
        &   & DDPM & 4.2k \\
        &   & iDDPM & 4.2k \\
        &   & PNDM & 4.2k \\
    \cmidrule{2-4}
        & \multirow{7}{*}{Text2Image} & LDM & 4.2k \\
        &   & SD-v1 & 4.2k \\
        &   & SD-v2 & 4.2k \\
        &   & VQ-Diffusion & 4.2k \\
        &   & IF & 1.0k \\
        &   & DALL·E 2 & 0.5k \\
        &   & Midjourney & 0.1k \\
    \midrule
    \multirow{4}{*}{ImageNet}  & Real & None & 5.0k \\
    \cmidrule{2-4}
        & Unconditional & ADM & 5.0k \\
    \cmidrule{2-4}
        & Text2Image  & SD-v1 & 5.0k \\
    \midrule
    \multirow{5}{*}{CelebA}  & Real & None & 4.2k \\
    \cmidrule{2-4}
        & \multirow{4}{*}{Text2Image}  & SD-v2 & 4.2k \\
        &   & IF & 1.0k \\
        &   & DALL·E 2 & 0.5k \\
        &   & Midjourney & 0.1k \\
    \bottomrule
\end{tabular}} 
\label{tab:df}
\end{table}
\begin{table}
\caption{The test dataset composition of GenImage.}
\centering
\resizebox{1.0\linewidth}{!}{
\begin{tabular}{lllr}
    \toprule
    Image Source & Category & Generator & \# Images \\
    \midrule
    \multirow{9}{*}{ImageNet}  & Real & None & 5.0k \\
    \cmidrule{2-4}
        & \multirow{2}{*}{Unconditional} & BigGAN & 6.0k \\
        &   & ADM & 6.0k \\
    \cmidrule{2-4}
        & \multirow{6}{*}{Text2Image} & SD-v1.4 & 6.0k \\
        &   & SD-v1.5 & 8.0k \\
        &   & VQ-Diffusion & 6.0k \\
        &   & GLIDE & 6.0k \\
        &   & wukong & 6.0k \\
        &   & Midjourney & 6.0k \\
    \bottomrule
\end{tabular}} 
\label{tab:gen}
\end{table}

\textbf{DiffusionForensics} is a collection of images generated by a variety of diffusion models, including ADM~\cite{dhariwal2021diffusion}, DDPM~\cite{ho2020denoising}, iDDPM~\cite{nichol2021improved}, PNDM~\cite{liu2022pseudo}, LDM~\cite{rombach2022high}, Stable Diffusions~\cite{rombach2022high}, VQ-Diffusion~\cite{gu2022vector}, IF~\cite{saharia2022photorealistic}, DALL·E 2~\cite{ramesh2022hierarchical} and Midjourney~\cite{midjourney2022}, designed for comprehensive experimentation. They can be broadly categorized into three splits based on their source: the LSUN-Bedroom, the ImageNet dataset, and the CelebA.

\textbf{GenImage} is a comprehensive dataset comprising 1.3 million images, generated using the latest state-of-the-art Diffusion and GAN models. This collection matches the number of real images in ImageNet, leveraging 1000 class labels from the dataset. The aim of \textbf{GenImage} is to tackle the challenge of detecting synthetic images produced by current top-tier generators. To achieve this, it employs eight distinct generative models, including BigGAN, GLIDE~\cite{nichol2021glide}, VQDM, Stable Diffusion v1.4, Stable Diffusion v1.5, ADM, Midjourney, and Wukong~\cite{wukong2023}.

\begin{figure*}[t]
    \centering
    \includegraphics[width=\linewidth]{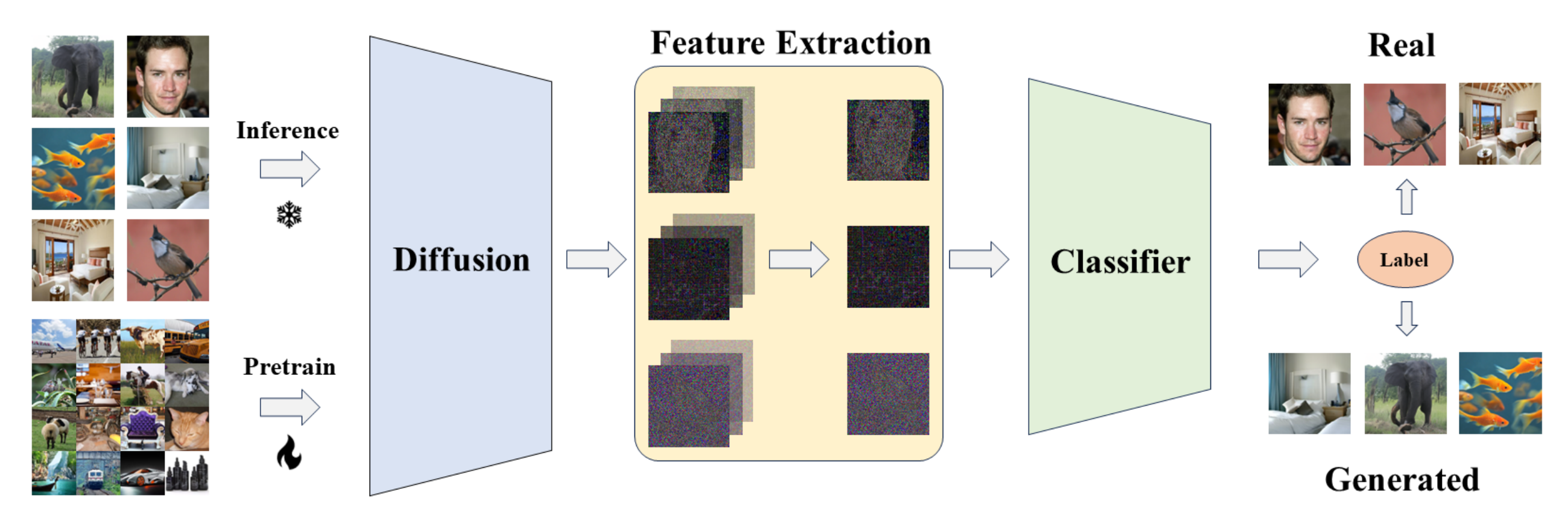}
    \caption{Overview of DNF.}
    \label{fig:overview}
\end{figure*}

\subsection{Baselines}
We will introduce  baselines in our experiments, \textbf{CNNDetection}~\cite{wang2020cnn}, \textbf{SBI}~\cite{shiohara2022detecting}, \textbf{Patchforensics}~\cite{chai2020makes}, \textbf{F3Net}~\cite{qian2020thinking}, and \textbf{DIRE}~\cite{wang2023dire}.

\textbf{CNNDetection} introduced a detection model trained to distinguish images generated by CNN-base models. However, this model exhibits strong generalization ability, meaning it can effectively detect images generated by various CNN models, not just the one it was trained on.

\textbf{SBI} is a method applied for DeepFake detection. It trains a universal generated image detector by blending fake source images with target images derived from a single original image. This approach enables the detector to learn and recognize synthetic images, regardless of the specific source or target images used in the blending process. 

\textbf{Patchforensics} utilizes a patch-wise classifier, which has been reported to outperform simple classifiers in detecting fake images. Instead of analyzing entire images, Patchforensics focuses on examining smaller patches within an image to identify inconsistencies or anomalies that indicate image manipulation or forgery. 

\textbf{F3Net} emphasizes the significance of frequency information in detecting generated images. By analyzing the frequency components of an image, F3Net can identify discrepancies or irregularities that are indicative of image tampering or generation. 

\textbf{DIRE} utilizes diffusion models to reconstruct images and uses the difference between the original image and the reconstructed image as the feature for classification. By comparing the differences between the features of real images and generated images, excellent performance in generated image detection can be achieved.

\section{Overview}

The overall DNF workflow is illustrated in Figure~\ref{fig:overview} to provide readers with a clearer understanding of our method.

\end{document}